\documentclass[10pt,twocolumn,letterpaper]{article}
\usepackage[final]{cvpr}
\usepackage[pagebackref=true,breaklinks=true,
letterpaper=true,colorlinks,citecolor=blue,
linkcolor=blue,bookmarks=false]{hyperref}
\usepackage{cite}
\usepackage{times}
\usepackage{epsfig}
\usepackage{graphicx}
\usepackage{amsmath}
\usepackage{amssymb}
\usepackage{enumitem}
\usepackage{rotating}

\usepackage{multirow}

\usepackage[capitalize]{cleveref}
\crefname{section}{Sec.}{Secs.}
\Crefname{section}{Section}{Sections}
\Crefname{table}{Table}{Tables}
\crefname{table}{Tab.}{Tabs.}

\def\cvprPaperID{14908} 
\def\confName{CVPR}
\def\confYear{2024}
\def\confYear{CVPR 2024}
\newcommand{\supp}[1]{}
\newcommand{\tmpdel}[1]{}

\begin{document}

\title{3D Points Splatting for Real-Time Dynamic Hand Reconstruction}

\author{Zheheng Jiang\textsuperscript{1}
~~~ Hossein Rahmani\textsuperscript{1}
~~~ Sue Black\textsuperscript{2} 
~~~ Bryan M. Williams\textsuperscript{1}\\
\textsuperscript{1}Lancaster University ~~
\textsuperscript{2}St John's College of the University of Oxford ~~ \\
{\tt\small \{z.jiang11,h.rahmani,b.williams6\}@lancaster.ac.uk, sue.black@sjc.ox.ac.uk}\\ 
}

\maketitle

\begin{abstract}
We present 3D Points Splatting Hand Reconstruction (3D-PSHR), a real-time and photo-realistic hand reconstruction approach. We propose a self-adaptive canonical points upsampling strategy to achieve high-resolution hand geometry representation. This is followed by a self-adaptive deformation that deforms the hand from the canonical space to the target pose, adapting to the dynamic changing of canonical points which, in contrast to the common practice of subdividing the MANO model, offers greater flexibility and results in improved geometry fitting. To model texture, we disentangle the appearance color into the intrinsic albedo and pose-aware shading, which are learned through a Context-Attention module. Moreover, our approach allows the geometric and the appearance models to be trained simultaneously in an end-to-end manner. We demonstrate that our method is capable of producing animatable, photorealistic and relightable hand reconstructions using multiple datasets, including monocular videos captured with handheld smartphones and large-scale multi-view videos featuring various hand poses. We also demonstrate that our approach achieves real-time rendering speeds while simultaneously maintaining superior performance compared to existing state-of-the-art methods.
\end{abstract}

\section{Introduction}
With the rapid development of virtual and augmented reality devices and technologies, there is a growing focus in the research community on achieving photo-realistic reconstructions of humans or specific human body parts \cite{cao2022authentic,grassal2022neural,jiang2022selfrecon,weng2022humannerf,zheng2023pointavatar}. Among these, human hands play a key role as the primary interface for interactions between humans and the digital world. The ability to create realistic animatable human hands holds great significance, with wide ranging applications in human computer interaction and entertainment.

Recent approaches for photo-realistic reconstruction of 3D human hands from videos rely on a 3D morphable MANO model \cite{qian2020html,kopanas2022neural,karunratanakul2023harp,chen2021model,jiang2023probabilistic,chen2023hand,guo2023handnerf,corona2022lisa}. These approaches often employ explicit mesh representations for hand shapes, followed by UV maps\cite{qian2020html,kopanas2022neural,karunratanakul2023harp}, vertex colours\cite{chen2021model,jiang2023probabilistic} or neural radiance
field (NeRF)\cite{chen2023hand,corona2022lisa,guo2023handnerf} for modeling hand appearance. However, existing UV map and vertex color-based methods, while efficient for rendering, suffer from limitations in appearance diversity imposed by their PCA-based texture parameter space \cite{qian2020html,kopanas2022neural} and the mesh resolution of the MANO model \cite{qian2020html,kopanas2022neural,chen2021model,jiang2023probabilistic}. Moreover, these approaches do not disentangle the 3D hand reconstruction into canonical and deformed spaces, overlooking the impact of pose variation on illumination and hand texture. 
Recent attempts using NeRF \cite{mildenhall2021nerf} have demonstrated remarkable success in representing hand geometry and appearance \cite{chen2023hand,corona2022lisa,guo2023handnerf}. 
 \begin{figure}
\begin{center}
\includegraphics[width=7.5cm]{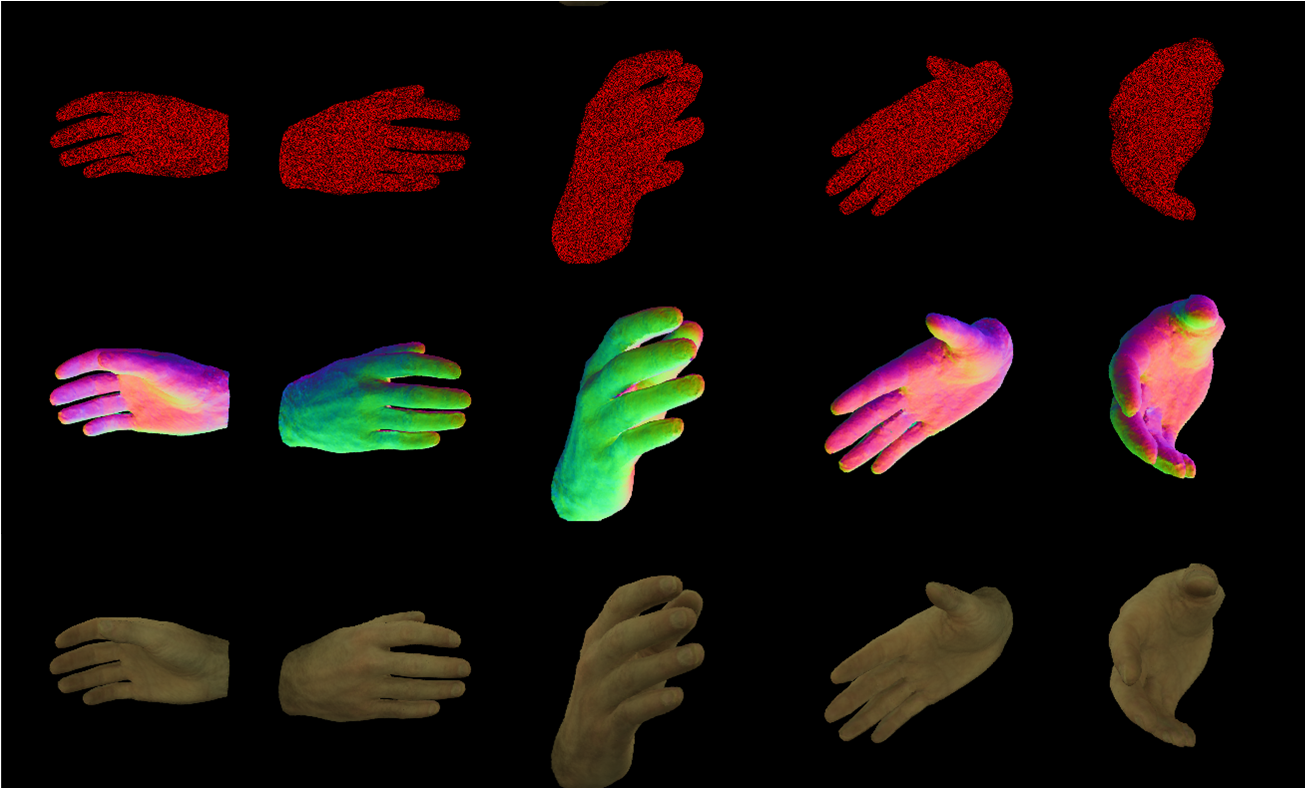}
\end{center}
\vspace{-6mm}
\caption{Example rendering by our free-pose 3D-PSHR, 
producing point clouds (top row), normal images (middle) and high-resolution 3D reconstruction with pose-aware shade (bottom).}
\vspace{-1.5em}
\label{fig:ablation_disentangled}
\end{figure}
Despite the superior performance of NeRF in rendering fidelity, these approaches suffer from significant training and rendering costs. This is due to the necessity of querying numerous points along the camera ray to render a single pixel, limiting their applicability for high-resolution point clouds. 
An overview of the existing \tmpdel{hand appearance} models is provided in Tab. \ref{tab:baselines}.

Driven by the above challenges, we propose a 3D points splatting approach designed to achieve real-time, realistic hand rendering.
To address limitations in mesh resolution, our method introduces a novel self-adaptive canonical points upsampling strategy instead of subdividing the MANO template. This enhancement allows the model to strategically allocate points in areas characterized by high curvature or complexity, resulting in a more accurate representation of hand geometry.
To ensure a smooth surface for the canonical points, we incorporate a smooth signed distance function with a regularization loss. Further, we introduce dynamic blendshapes to facilitate pose-aware deformation, enabling the disentanglement of RGB color into intrinsic albedo and pose-aware shading. Leveraging a context-attention module enhances the model's ability to capture contextual information associated with each point, contributing to improved rendering fidelity.

A key feature of our model lies in its ability to adapt to changes in hand pose. Normal deformation is employed to effectively capture the impact of pose variation on each point, enhancing the model's pose-aware shading capabilities. Additionally, a 3D points splatting approach is introduced to enable our approach to achieve real-time speed while maintaining superior performance.
To demonstrate the versatility of our proposed model, we showcase its adaptability to different training scenarios. Whether trained with multi-view or single-view RGB video, our model exhibits robust performance. Further, we highlight its capability for relighting and rendering with novel poses from various viewing directions, underscoring its potential for applications in diverse contexts.

The contributions of this paper are as follows:
\textbf{1)} To overcome efficiency challenges while increasing accuracy, we propose a 3D Points Splatting (3PS) framework for photo-realistic hand reconstruction, which shows state-of-the-art visual quality while being considerably more efficient than existing neural hand rendering methods. \textbf{2)} We propose a two-fold approach for achieving high-resolution fit to hand geometry and pose through our self-adaptive canonical points upsampling (SAU) strategy and self-adaptive deformation (SAD), facilitating pose-free hand reconstruction. 
\textbf{3)} We disentangle the appearance color into the intrinsic albedo and the normal deformation based pose-aware shading, both of which are learned through a Context-Attention module. \textbf{4)} Our extensive experiments show that our approach achieves both real-time rendering speed and superior performance to the existing state-of-the-art.
%
%
%
\begin{table}
\begin{center}
\footnotesize
\caption{Overview of existing hand appearance models}
\vspace{-2mm}
\begin{tabular}{l|p{1.8cm}|p{1cm}|p{0.6cm}|p{1.1cm}}
\hline
 & Appearance Representation& Canonical Space &Pose driven&Efficient Rendering\\
\hline
HTML\cite{qian2020html}&UV map& \hfil 
X &\hfil X& \hfil \checkmark \\
NIMBLE\cite{li2022nimble}&UV map& \hfil 
X &\hfil X& \hfil \checkmark \\
HARP\cite{karunratanakul2023harp}&UV + normal& \hfil 
X &\hfil X& \hfil \checkmark \\
LISA\cite{corona2022lisa}&NeRF& \hfil 
\checkmark &\hfil X& \hfil X \\
HandNeRF\cite{guo2023handnerf}&NeRF& \hfil 
\checkmark &\hfil \checkmark& \hfil X \\
HandAvatar\cite{chen2023hand}&NeRF& \hfil 
\checkmark &\hfil \checkmark& \hfil X \\
$S^2$HAND\cite{chen2021model}&Vertex color& \hfil 
X &\hfil X& \hfil \checkmark \\
AMVUR\cite{jiang2023probabilistic}&Vertex color& \hfil 
X &\hfil X& \hfil \checkmark \\
ours&Points color& \hfil 
\checkmark &\hfil \checkmark& \hfil \checkmark \\
\hline
\end{tabular}
\label{tab:baselines}
\end{center}
\vspace{-1.5em}
\end{table}
\section{Related Work}
\subsection{Hand Geometry Representation}
Parametric meshes have become highly popular for modeling articulated objects such as human body and hands. The MANO hand model \cite{romero2017embodied} is a commonly used representation that employs pose and shape parameters, transforming them into a hand mesh through predetermined linear blend skinning.
However, the MANO model can only provides a low-resolution mesh with 778 vertices. Several recent works \cite{chen2023hand,karunratanakul2023harp} address this limitation by increasing the number of vertices through subdivision of the MANO template mesh at edge midpoints. Despite this improvement, these approaches maintain a fixed and predetermined number of vertices throughout training, restricting their adaptability to various hand geometries. Implicit geometry is another popular trend in geometry representation \cite{alldieck2021imghum, mihajlovic2021leap, karunratanakul2020grasping, zheng2022imface}, offering enhanced flexibility and continuity compared to explicit hand geometry. Lisa \cite{corona2022lisa} is the first method to learn an implicit hand geometry, although it relies on multi-view image capturing setup. Different from the above approaches, we propose a dynamic implicit hand geometry, where the canonical hand representation and implicit deformation undergo iterative updates with the increasing number of points. Moreover, our method accommodates both multi-view image capturing setup and monocular setup.
\subsection{Appearance Representation}
Traditional methods for hand appearance modeling typically involve learning PCA-based UV textures \cite{qian2020html,li2022nimble} from extensive 3D scans of hands in various poses. However, these approaches face limitations due to their reliance on a linear appearance space and small training set, making them unsuitable for generating personalized hand avatars from multi-view or monocular videos. HARP \cite{karunratanakul2023harp} introduces explicit albedo and normal UV maps to represent hand appearance on top of the MANO model, more effectively capturing personalized and animatable hands compared to PCA-based textures. However, HARP optimizes UV texture maps without utilizing a learning model, neglecting the impact of pose on illumination and hand texture.
$S^2$hand \cite{chen2021model} and AMVUR \cite{jiang2023probabilistic} focus on estimating vertex colors using MLP or self-attention, but their rendering quality is constrained by the sparsity of the MANO mesh. 
The recent success of neural rendering techniques has led to approaches \cite{corona2022lisa,deng2022gram,jiang2022neuman,liu2021neural,qiao2023dynamic,guo2023handnerf,mildenhall2021nerf,weng2022humannerf} that represent appearance in a neural radiance field (NeRF) \cite{mildenhall2021nerf}. Lisa \cite{corona2022lisa} employs NeRF to learn hand appearance by integrating volume rendering with a hand geometric prior. 
HandNeRF \cite{guo2023handnerf} combines a pose-conditioned deformation field and a neural radiance field to reconstruct the photorealistic appearance of single or interacting hands from multi-view images. In HandAvatar \cite{chen2023hand}, hand appearance is disentangled into albedo and an illumination field, followed by a volume rendering process to generate realistic hand texture. However, training these models can be computationally expensive and time-consuming.
%
Recently, PointNeRF \cite{xu2022point} combined point cloud representations with volume rendering, achieving faster training speed. However, it relies on an off-the-shelf multi-view stereo method for generating point geometry and requires keeping the points fixed during appearance optimization. 
In contrast, our method accommodates both multi-view and monocular setups,  thanks to our self-adaptive canonical hand representation \cref{sec:hand_representation} and self-adaptive deformation \cref{sec:deformation}. Further, our approach allows simultaneous optimization of point geometry and appearance in an end-to-end manner.
\subsection{Point-based Rendering}
Point-based neural rendering has recently gained significant attention due to its efficiency and flexibility. The most straightforward approach involves rendering points as one-pixel splats \cite{grossman1998point}. However, this suffers from issues such as holes and discontinuities. Studies in high-quality point-based rendering have addressed these challenges by employing "splatting" point primitives larger than a single pixel \cite{botsch2005high, pfister2000surfels, ren2002object, zwicker2001surface}. Recently, there has been a growing interest in differentiable point-based rendering techniques for scene representation \cite{wiles2020synsin,yifan2019differentiable}. 
Kerbl et al. \cite{kerbl20233d} introduced a differentiable point-based splatting technique that projects 3D Gaussians into 2D splats, enabling real-time rendering of novel views. However, these methods are primarily designed for static scenes, limiting their applicability to free-pose animation and rendering. The human hand is highly articulated with large and complex deformations, posing challenges for the application of point rendering techniques. A key challenge in developing animatable and realistic hand rendering lies in effectively modeling these deformations. The large variation in hand pose makes it challenging to accurately model hand texture, affecting both illumination and the overall appearance of the hand.
To address these issues, we introduce a normal-based deformation model, which not only enables point clouds to deform based on the hand pose but also captures the impact of pose variations on the appearance changes of individual points. Our work represents the pioneering effort in the domain of real-time, photo-realistic hand reconstruction using a 3D Points Splatting approach.

\section{Proposed Model}
\begin{figure*}
\begin{center}
\includegraphics[width=15cm]{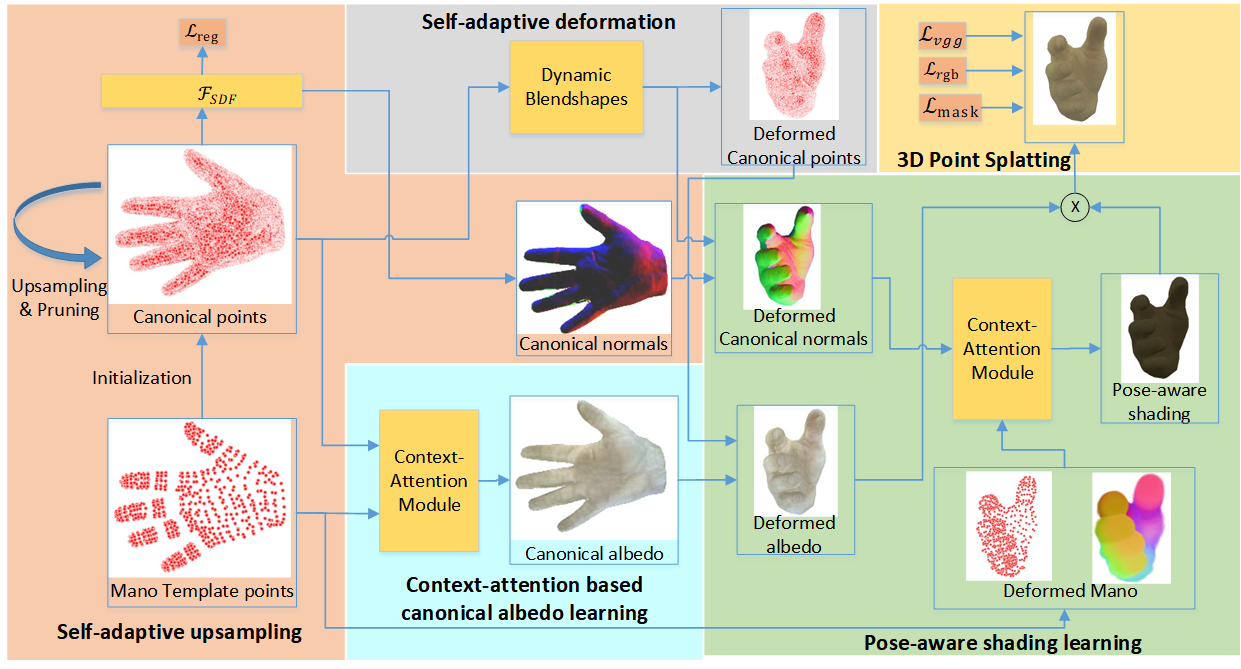}
\end{center}
\vspace{-6mm}
\caption{ Overview of our proposed method. The high mesh resolution of the canonical hand is achieved by using self-adaptive canonical points upsampling strategy, followed by a self-adaptive deformation process to obtain deformed canonical points. We disentangle the RGB color into a intrinsic
albedo and a pose-aware shading component. The albedo value is learned through a context-attention module by taking the point coordinates as input. Similar, another context-attention module is applied to obtain the pose-aware shading value from the deformed canonical normals. The canonical point normals can be computed as the spatial gradient of a signed distance function. }
\vspace{-1.5em}
\label{fig:framework}
\end{figure*}

Given a multi-view/single-view RGB video capturing a hand in various poses, our objective is to reconstruct a photo-realistic, deformable hand model capable of generalizing to novel poses from previously unseen viewing angles. Figure \ref{fig:framework} provides an overview of our proposed model, encompassing both our geometric and appearance models. The geometric model comprises two primary components: Self-adaptive canonical hand representation (Sec. \ref{sec:hand_representation}) and Self-adaptive deformation (Sec. \ref{sec:deformation}). Our appearance model includes Context-attention based canonical albedo learning (Sec. \ref{sec:albedo}) and Pose-aware shading learning (Sec. \ref{sec:shading}). Finally, we introduce a differentiable point rendering method (Sec. \ref{sec:rendering}), serving the dual purpose of generating 2D hand images from 3D hand models and optimizing parameters within our geometric and appearance models.

\subsection{Self-adaptive canonical hand representation}\label{sec:hand_representation}
To achieve high canonical hand mesh resolution, existing approaches \cite{chen2023hand,karunratanakul2023harp,corona2022lisa} typically subdivide the MANO template mesh along edges to increase the number of points, limiting their adaptability to various hand geometries. In contrast, we propose a self-adaptive canonical points upsampling (SAU) strategy, which dynamically adjusts the density of canonical points in response to variations in hand pose. Simultaneously trained with our appearance model, SAU allows the model to concentrate more points in regions with high curvature or complexity, resulting in a more accurate representation of the hand's geometry.


To represent a canonical hand, we define a set of learnable points $P_{C} = \left\{p_{C}^{i}\right\}_{i=1}^{N_{C}}$, where $C$ denotes canonical space, $N_{C}$ is the total number of canonical points, and $p_{C}^{i}\in\mathbb{R}^3$ represents the optimizable point coordinates for the $i$-th point in canonical space. Note that $N_{C}$ is not fixed; it dynamically adjusts and increases after every \textit{e} epochs, where \textit{e} is empirically set to 5.


 \begin{figure}
\begin{center}
\includegraphics[width=7cm]{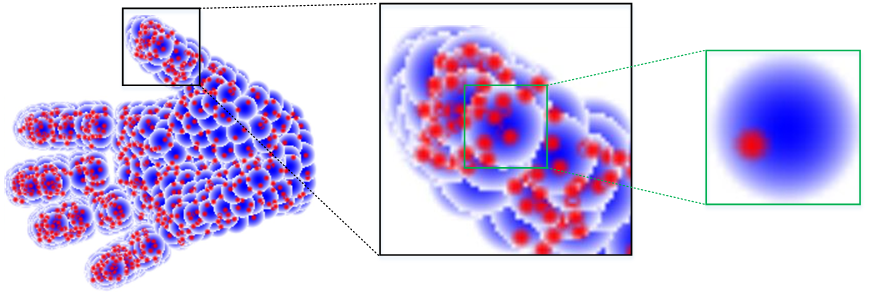}
\end{center}
\vspace{-5mm}
\caption{Self-adaptive canonical points upsampling (SAU). This example shows newly upsampled canonical points (red) within spherical regions (blue) centered on the canonical points generated in the previous upsampling step.\tmpdel{ In each upsampling step, we randomly upsampled one point within each sphere.}}
\vspace{-1.5em}
\label{fig:sampling}
\end{figure}

In SAU, we first initialize our canonical points with the MANO template. Then periodically increase $N_C$ through upsampling within spherical regions centered on the canonical points generated in the preceding upsampling step, as shown in \cref{fig:sampling}, while reducing the radius of these points. 
However, such a sampling strategy struggles to produce a smooth surface. To alleviate this issue, we propose to regularize the implicit surface of the canonical points by aligning their locations with the zero-level set. To achieve this, we train an MLP to approximate a signed distance function (SDF) $\mathcal{F}_{SDF}(\cdot)$ by taking the coordinates of each canonical point as input. Then, our regularization loss is defined as:
\begin{equation}
\small
\begin{split}
\mathcal{L}_{reg}=&\lambda_{SDF}\sum_{i=1}^{N_{C}}\left \|\mathcal{F}_{SDF}(p_{C}^{i})\right \|^{2} + \\&\lambda_{eik}\sum_{p\in \left\{P_{C}, \Omega\right\}}(\left \|\bigtriangledown_{p}\mathcal{F}_{SDF}(p)\right \|-1)^{2}  
\end{split}
\end{equation}
where $\lambda_{SDF}$ and $\lambda_{eik}$ are loss weights. The second term is the Eikonal term \cite{gropp2020implicit}, which ensures the MLP approximates a signed distance function. The set $\Omega$ comprises points randomly sampled around the canonical points\tmpdel{, which is only employed in the calculation of the Eikonal term}. At the end of each epoch we prune outlier points that fall outside the foreground boundary of the hand, which is generated by projecting the MANO mesh.

\subsection{Self-adaptive deformation}\label{sec:deformation}
Inspired by \cite{romero2022embodied}, we  deform the canonical points via blendshapes. Rather than relying on static, predetermined blendshapes, we propose dynamic blendshapes that adapt to the growing number of canonical points.
We define shape blendshapes $\mathcal{B}_{s}(\phi;\mathcal{S})$, pose blendshapes $\mathcal{B}_{p}(\theta;\mathcal{P})$ and linear blend skinning weights $W \in \mathbb{R}^{N_{C}\times N_{j}}$. Here, $N_{j}$ denotes the number of bones, $\phi$ represents a 10-dimensional shape parameter, and $\theta$ is a $N_{j}\times 3$ dimension pose parameter indicating joint rotations in axis-angle form. $\mathcal{B}_{s}$ and $\mathcal{B}_{p}$ calculate additive shape and pose offsets using the corrective blendshape bases $\mathcal{S}\in \mathbb{R}^{N_{C}\times 3 \times 10}$ and $\mathcal{P}\in \mathbb{R}^{N_{C}\times 3 \times 135}$.\tmpdel{ As our SAU strategy is applied, the value of $N_{C}$ undergoes changes throughout the training process.} 

To determine $\mathcal{S}$, $\mathcal{P}$, and $W$ for \tmpdel{all points in} the newly generated canonical points, we identify the nearest vertices $\mathcal{N}$ for each canonical point within the MANO template. This allows us to obtain $\mathcal{S}=\mathcal{S}_{M}(\mathcal{N}), \mathcal{P}=\mathcal{P}_{M}(\mathcal{N}),\mathcal{W}=\mathcal{W}_{M}(\mathcal{N})$. $\mathcal{S}_{M},\mathcal{P}_{M}$, and $\mathcal{W}_{M}$ are blendshape bases and skinning weights defined in the MANO template. The point coordinates of each canonical point in the deformed space $D$ is computed as:
\begin{equation}
\small
\begin{split}
p_{D}^{i}=\sum_{j=1}^{N_{j}} (w^{i,j})T_{j}\mathcal{H}(p_{C}^{i}+\mathcal{B}_{s}^{i}(\phi;\mathcal{S}^{i}) + \mathcal{B}_{p}^{i}(\theta;\mathcal{P}^{i}))
\end{split}
\label{eq:deformation}
\end{equation}
where $T_{j} \in \mathbb{R}^{4\times 3}$ is the rigid transformation applied on the bone j, $w_{i,j}$ is the $(i, j)$ entry of W. $\mathcal{H}$ denotes the homogeneous coordinate transformation. We obtain deformed canonical points, allowing us to calculate the hand appearance in the deformed space, as shown in \cref{fig:framework}. 

\tmpdel{Next, we present our appearance model consisting of Context-attention based canonical albedo learning (\cref{sec:albedo}) and Pose-aware shading learning (\cref{sec:shading}).
}
\subsection{Context-attention based canonical albedo learning}\label{sec:albedo}
Canonical Albedo describes the intrinsic appearance of the hand in the canonical space, unaffected by varying pose and illumination. An MLP is typically employed to predict the albedo value for computational efficiency \cite{chen2023hand,karunratanakul2023harp}. However, the MLP has a fixed local receptive field, limiting its ability to effectively explore contextual information associated with points. This limitation leads to a loss of fine details in the reconstructed 3D hand. Leveraging the high computational efficiency of our rendering approach (\cref{sec:rendering}), we integrate a more sophisticated yet efficient albedo learning architecture to encode such contextual information. We associate each canonical point, generated via our SAU, with the MANO template through an attention mechanism. This involves initially mapping the coordinates of MANO template points $P_{M}$ and canonical points $P_{C}$ to a hidden space using an MLP. Subsequently, the vertex-to-vertex similarity between canonical points and MANO template points are calculated as:
\begin{equation}
\small
\begin{split}
S_{cross} = softmax\left(\frac{q_{cross}k_{cross}^T}{\sqrt{d_{cross}}}\right)
\end{split}
\label{eq:cross}
\end{equation}
where $q_{cross} = W_{q}\mathcal{F}_{MLP}(P_{C})$ and $k_{cross} = W_{k}\mathcal{F}_{MLP}(P_{M})$ denote the query and key embedding. $d_{cross}$ denotes the feature dimension of the key $k_{cross}$. 
Point features are then refined by aggregating this contextual information: 
$\mathcal{F}_{cross}= W_{v}\mathcal{F}_{MLP}(P_{M})S_{cross}$, where $W_{q},W_{k},W_{v}$ are trainable parameters applied for query, key and value embedding. 
Finally, a further MLP is used to map the results to the albedo colors. The architecture of our context-attention module is shown in Suppl. file. \tmpdel{Given the deformed canonical points, generating deformed albedo is straightforward. However, it is crucial to recognize that albedo is independent of hand pose, while illumination and hand texture vary with changes in hand pose. To address this, we introduce a pose-aware shading learning method,  which we present in detail next.}

\subsection{Pose-aware shading learning}\label{sec:shading}
It is crucial to recognize that albedo is independent of hand pose, while illumination and hand texture vary with changes in hand pose. To address this, we introduce a pose-aware shading learning method. 
As shown in \cref{fig:framework}, incorporation of dynamic blendshapes and canonical normals facilitates the derivation of deformed normals, effectively capturing pose changes. Changes in hand pose can result in variations in hand texture and shading, influenced by the diverse ways light interacts with the hand surface. Therefore, understanding and modeling of these pose-related changes are essential for achieving accurate and realistic rendering of hand appearance. However, most previous works \cite{qian2020html, kopanas2022neural, chen2021model, jiang2023probabilistic, corona2022lisa} have not specifically addressed this issue. 
While attempts have been made in \cite{chen2023hand,guo2023handnerf} to address this challenge, these approaches rely on explicit hand geometry with NeRF, making them unsuitable for direct integration into our method. Furthermore, the utilization of ray tracing in NeRF necessitates evaluation of the radiance field at numerous points along each ray, incurring considerable computational cost. Our approach addresses this challenge through the integration of normal deformation. 

To estimate the normal of each canonical point, we calculate the spatial gradient $\mathcal{F}_{SDF}(\cdot)$ of the SDF as
\begin{equation}
\small
n_{C}^{i}=\bigtriangledown_{p_{C}^{i}}\mathcal{F}_{SDF}(p_{C}^{i}).
\end{equation}

To compute the normal deformation, the normalized gradient of the $\mathcal{F}_{SDF}(p_{C}^{i})$, w.r.t. the location of the deformed point is computed as:
\begin{equation}
\small
\frac{\partial \mathcal{F}_{SDF}(p_{C}^{i})}{\partial p_{D}^{i}}=\frac{\partial \mathcal{F}_{SDF}(p_{C}^{i})}{\partial p_{C}^{i}}\frac{\partial p_{C}^{i}}{\partial p_{D}^{i}}=n_{C}^{i}(\frac{\partial p_{D}^{i}}{\partial p_{C}^{i}})^{-1},
\label{eq:normaldef}
\end{equation}
where $(\frac{\partial p_{D}^{i}}{\partial p_{C}^{i}})^{-1}$ can be calculated using the inverse of the Jacobian of \cref{eq:deformation}.
\tmpdel{\begin{equation}
\small
(\frac{\partial p_{D}^{i}}{\partial p_{C}^{i}})^{-1}= \begin{bmatrix} 
   \frac{\partial p_{D}^{i,x}}{\partial p_{C}^{i,x}} & \frac{\partial p_{D}^{i,y}}{\partial p_{C}^{i,x}}  &  \frac{\partial p_{D}^{i,z}}{\partial p_{C}^{i,x}}\\
     \frac{\partial p_{D}^{i,x}}{\partial p_{C}^{i,y}} &  \frac{\partial p_{D}^{i,y}}{\partial p_{C}^{i,y}} &  \frac{\partial p_{D}^{i,z}}{\partial p_{C}^{i,y}}\\
    \frac{\partial p_{D}^{i,z}}{\partial p_{C}^{i,x}} & \frac{\partial p_{D}^{i,z}}{\partial p_{C}^{i,y}}  & \frac{\partial p_{D}^{i,z}}{\partial p_{C}^{i,z}} 
    \end{bmatrix}^{-1}
\end{equation}
where $(x,y,z)$ indicate the 3D coordinates.}
Similar to our Context-attention based canonical albedo learning, cross-attention is employed to map the gradients capturing the normal deformation of canonical points to shading values. This involves initially mapping the normal deformations of the canonical points and MANO template points to a hidden space using an MLP. Subsequently, the similarity between the normal deformation of canonical points and MANO template points are calculated as:
\begin{equation}
\small
\begin{split}
S_{cross}' = softmax\left(\frac{q_{cross}'k_{cross}'^T}{\sqrt{d_{cross}'}}\right)
\end{split}
\label{eq:cross}
\end{equation}
where $q_{cross}' = W_{q}'\mathcal{F}_{MLP}(\mathcal{D}_C)$ and $k_{cross}' = W_{k}'\mathcal{F}_{MLP}(\mathcal{D}_M)$ denote the query and key embedding. $\mathcal{D}_C\in \mathbb{R}^{N_{C}\times 3}$ and $\mathcal{D}_{M}\in \mathbb{R}^{N_{M}\times 3}$ denote the normal deformation of canonical points and MANO template points, respectively. $N_{M}$ is the number of points in the MANO template. The normal deformation of each canonical point in $\mathcal{D}_C$ is calculated using \cref{eq:normaldef}. Similarly, the normal deformation of each MANO template point in $\mathcal{D}_M$ is computed using $n_{M}^{i}(\frac{\partial p_{D,M}^{i}}{\partial p_{M}^{i}})^{-1}$, where $n_{M}^{i}$, $p_{D,M}^{i}$, and $p_{M}^{i}$ represent the normal, deformed point, and canonical point of the MANO model, respectively. Point features are then refined: $\mathcal{F}_{cross}'= W_{v}'\mathcal{F}_{MLP}'(\mathcal{D}_{M})S_{cross}'$, where $W_{q}',W_{k}',W_{v}'$ are trainable parameters for query, key and value embedding. An MLP is then used to map the results to shading values.
Finally, the color of each point in the deformed space is determined by the element-wise product between the albedo colors and shading values. As demonstrated in \cref{sec:relighting}, our approach enables relighting using the Phoen model \cite{phong1998illumination} and the self-shadow model \cite{karunratanakul2023harp}.

\subsection{Differentiable Point Rendering} \label{sec:rendering}
To optimize all parameters of our geometric and appearance models, a differentiable rendering approach is crucial. This not only generates 2D hand images from the 3D model but also aids in computing gradients for optimization. Recently, NeRF based approaches are widely used for image rendering \cite{chen2023hand,corona2022lisa}. NeRF\tmpdel{, as a volume rendering algorithm,} demands expensive random sampling along camera rays to represent a continuous scene\tmpdel{, particularly challenging for high-resolution point clouds.} This can result in a large number of queries and significant computational expense.  In contrast, we propose a differentiable point rendering method that is efficient and flexible enough to approximate geometry by optimizing opacity and positions, while circumventing the limitations associated with volumetric representations.  

In our point-based rendering approach, each point is represented as a sphere with a periodically reduced radius by a factor of $1/\sqrt{2}$ after each upsampling step. The rendering process involves splatting each point as a 2D circle onto the image and integrating $N_{z}$ ordered points in the z-buffer that overlap with the pixel. The color of the image pixel from the overlapping points in the z-buffer is computed as:
\begin{equation}
\small
c_{pix}=\sum_{i \in N_{z}}c_{i}a_{i}\prod_{j=1}^{i-1}(1-a_{j})
\end{equation}
where $c_{i}$ denotes the color value of $i$-th point in the z-buffer, and $a_{i}$ is calculated as $a=1-\frac{d^{2}}{r^{2}}$, where $d$ denotes the distance from the overlapping points in the z-buffer to the pixel, and $r$ denotes the splatting radius. We create the rendered image $C$ by collecting all $c_{pix}$ within an image. 

\subsection{Training and inference}
To optimize our model, the primary loss is the rgb loss, calculated as the difference between the rendered image $C$ and the ground truth image $\hat{C}$:  
$
\mathcal{L}_{rgb}=\|C - \hat{C}\|.
$
To capture the perceptual difference between
the rendered images and the ground truth, we compare the features extracted from VGG, and adopt a VGG feature loss\cite{ledig2017photo}: 
$
\mathcal{L}_{vgg}=\|\mathcal{F}_{vgg}(C) - \mathcal{F}_{vgg}(\hat{C})\|.
$
To ensure alignment between the rendered silhouette and the hand mask while adhering to 3D mesh constraints, we incorporate the mask loss: 
$
\mathcal{L}_{mask}=\|M - \hat{M}\|,
$
where $M$ and $\hat{M}$ denote the predicted and ground truth hand masks. Our final loss is: 
\begin{equation}
\small
\mathcal{L}=\lambda_{rgb}\mathcal{L}_{rgb} + \lambda_{vgg}\mathcal{L}_{vgg}+\lambda_{mask}\mathcal{L}_{mask} + \lambda_{reg}\mathcal{L}_{reg},
\end{equation}
where $\lambda_{rgb}$, $\lambda_{vgg}$, $\lambda_{mask}$ and $\lambda_{reg}$ are assigned weights\tmpdel{ assigned to $\mathcal{L}_{rgb}$, $\mathcal{L}_{vgg}$, $\mathcal{L}_{mask}$ and $\mathcal{L}_{reg}$, respectively}. 
\tmpdel{
It is important to highlight that, c}Constrained by memory limitations, NeRF-based approaches utilize patch-based training, leading to decreased training speed. One advantage of our point-based rendering lies in its capacity to process batches of images at each training step. 
For inference, our approach can generate photo-realistic rendered hand images given any hand pose and view direction. Further,  \cite{phong1998illumination} and the self-shadow model \cite{karunratanakul2023harp} can be integrated with our approach to relight hand images.

\section{Experiments}
\noindent \textbf{Datasets:}
We compare our approach with SOTA methods on InterHand2.6M\cite{moon2020interhand2}, which contains large-scale multi-view sequences of various hand poses. For fair comparison, we follow \cite{chen2023hand} to choose training data from the ROM04\underline{~}RT\underline{~}Occlusion sequence and validation data
from the ROM03\underline{~}RT\underline{~}No\underline{~}Occlusion sequence. The number of samples in the training and validation sets is shown in \cref{tab:data}. In each frame, we extract the hand region using an adjusted annotated bounding box,\tmpdel{, adjusting it to achieve a square shape with a $1.3$ times expansion factor. Subsequently, the hand region is} cropped and resized to $256\times 256$ resolution. We utilize the Hand Appearance dataset \cite{karunratanakul2023harp}, comprising single-view hand sequences captured by a hand-held smartphone camera\tmpdel{ pointing at a right hand}. \tmpdel{Unlike the InterHand2.6M dataset, t}This dataset includes both hand and arm in each frame. We follow \cite{karunratanakul2023harp}, resizing all frames to $448 \times 448$ pixels and dividing them into 750 training frames and 600 testing frames. \tmpdel{Note that t}The evaluation data comprises novel poses and viewing directions not seen during training.

\begin{table}
\begin{center}
\footnotesize
\caption{Data amount for training and validation sets}
\begin{tabular}{l|p{2cm}|c}
\hline
Capture & Training set& Testing set\\
\hline
test/Capture0 & 11,757 & 194\\
test/Capture1 & 18,474 & 232\\
val/Capture0 & 18,340 & 184\\
\hline
\end{tabular}
\label{tab:data}
\end{center}
\vspace{-1.5em}
\end{table}

\noindent \textbf{Baselines:}
\tmpdel{An overview of the existing baselines for generating hand avatars from RGB video data is provided in \cref{tab:baselines}. }We compare our method with UVmap-based, NeRF-based and Vertex-based approaches\tmpdel{, which are commonly used techniques for creating hand avatars}. LISA\cite{corona2022lisa} and HandNeRF\cite{guo2023handnerf} are excluded from the comparison due to the unavailability of their source code and datasets.

\noindent \textbf{Metrics:}
Following previous work \cite{chen2023hand,corona2022lisa}, we evaluate rendering quality with the silhouette intersection-over union (IoU), peak signal-to-noise ratio (PSNR), structural similarity index (SSIM), and learned perceptual image patch similarity (LPIPS). \tmpdel{To calculate IOU, w}We use the mask generated by HandAvatar as ground truth for the InterHand2.6M dataset. We use the segmentation tool employed by HARP\cite{karunratanakul2023harp} to obtain ground truth masks for the Hand Appearance dataset.

\noindent \textbf{Implementation:}
Our approach is implemented with PyTorch\cite{paszke2019pytorch}. 
We set $\lambda_{rgb} = 1$, $\lambda_{mask} = 1$, $\lambda_{vgg} = 0.1$, $\lambda_{reg} = 1$, $\lambda_{SDF} = 1$ and $\lambda_{eik} = 0.1$. 
We train our model using an Adam optimizer\cite{kingma2014adam} for 100 epochs, with a learning rate of $\mu=1e^{-4}$. We optimize our geometric model for 35 epochs and keep it fixed for the remaining of the training. \tmpdel{Concurrently, our appearance model undergoes continuous optimization throughout the entire training process. }In the multi-view and single-view experiments, we shuffle the training dataset and use a batch size of 16\tmpdel{ for training}. \tmpdel{All experiments were performed using a Dell Precision 5820 Workstation with a single NVidia GeForce RTX 4090 Ti GPU, 32GB RAM and an Intel(R) Xeon(R) W-2245 CPU.} Our source code and model weights will be made publicly available.
\begin{table*}
\begin{center}
\footnotesize
\caption{Appearance reconstruction results on the Inter-Hand2.6M dataset}
\begin{tabular}{l||p{0.8cm}p{0.8cm}p{0.8cm}p{0.8cm}||p{0.8cm}p{0.8cm}p{0.8cm}p{0.8cm}||p{0.8cm}p{0.8cm}p{0.8cm}p{0.8cm}}
\hline
\multirow{2}{*}{Method}
 & \multicolumn{4}{c||}{\textit{test/Capture0}}& \multicolumn{4}{c||}{\textit{test/Capture1}} &\multicolumn{4}{c}{\textit{val/Capture0}}\\
 &IOU$\uparrow$&LPIPS$\downarrow$&PSNR$\uparrow$&SSIM$\uparrow$&IOU$\uparrow$&LPIPS$\downarrow$&PSNR$\uparrow$&SSIM$\uparrow$&IOU$\uparrow$&LPIPS$\downarrow$&PSNR$\uparrow$&SSIM$\uparrow$\\
\hline
HTML\cite{qian2020html}&0.922&0.181&24.23&0.859&0.921&0.173&23.11&0.853&0.915&0.186&23.41&0.851\\
SelfRecon\cite{jiang2022selfrecon}&\centering/&0.142& 26.38 &0.879&/& 0.139& 25.18& 0.876&/& 0.149& 25.78& 0.869\\
HumanNeRF\cite{jiang2022selfrecon}&\centering/&0.115& 27.64 &0.884&/& 0.118& 26.31& 0.880&/& 0.119& 27.80& 0.882\\
$S^2$HAND\cite{chen2021model}&0.924&0.141&26.84&0.881&0.922&0.135&25.81&0.879&0.916&0.146&25.94&0.877\\
AMVUR\cite{jiang2023probabilistic}&0.937&0.135&27.21&0.886&0.932&0.126&25.43&0.884&0.928&0.132&27.43&0.885\\
HandAvatar\cite{chen2023hand}&0.924&0.104& 28.23 &0.894&0.922& 0.108& 26.56& 0.890&0.919& 0.106& 28.04& 0.890\\
ours&\textbf{0.946}&\textbf{0.078}&\textbf{30.93}&\textbf{0.934}&\textbf{0.948}&\textbf{0.089}&\textbf{29.23}&\textbf{0.913}&\textbf{0.938}&\textbf{0.092}&\textbf{29.40}&\textbf{0.910}\\
\hline
\end{tabular}
\label{tab:SOTA_results}
\end{center}
\vspace{-1.5em}
\end{table*}
\subsection{Comparison with SOTA Methods}
 \begin{figure}
\begin{center}
\includegraphics[width=7.5cm]{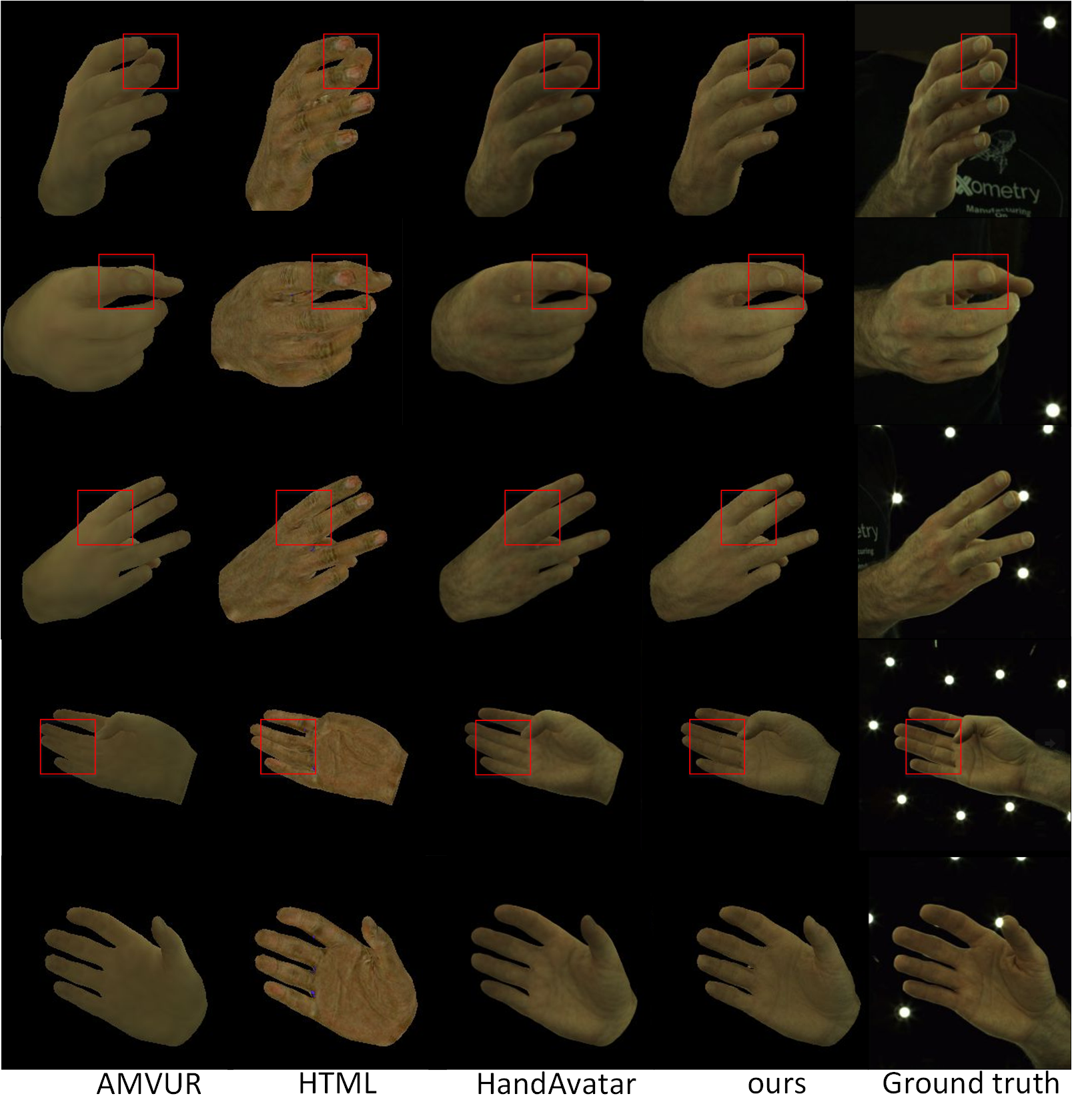}
\end{center}
\vspace{-6mm}
\caption{Comparison with SOTA Methods on InterHand2.6M}
\vspace{-1.5em}
\label{fig:InterHand_comp}
\end{figure}
 \begin{figure}
\begin{center}
\includegraphics[width=8cm]{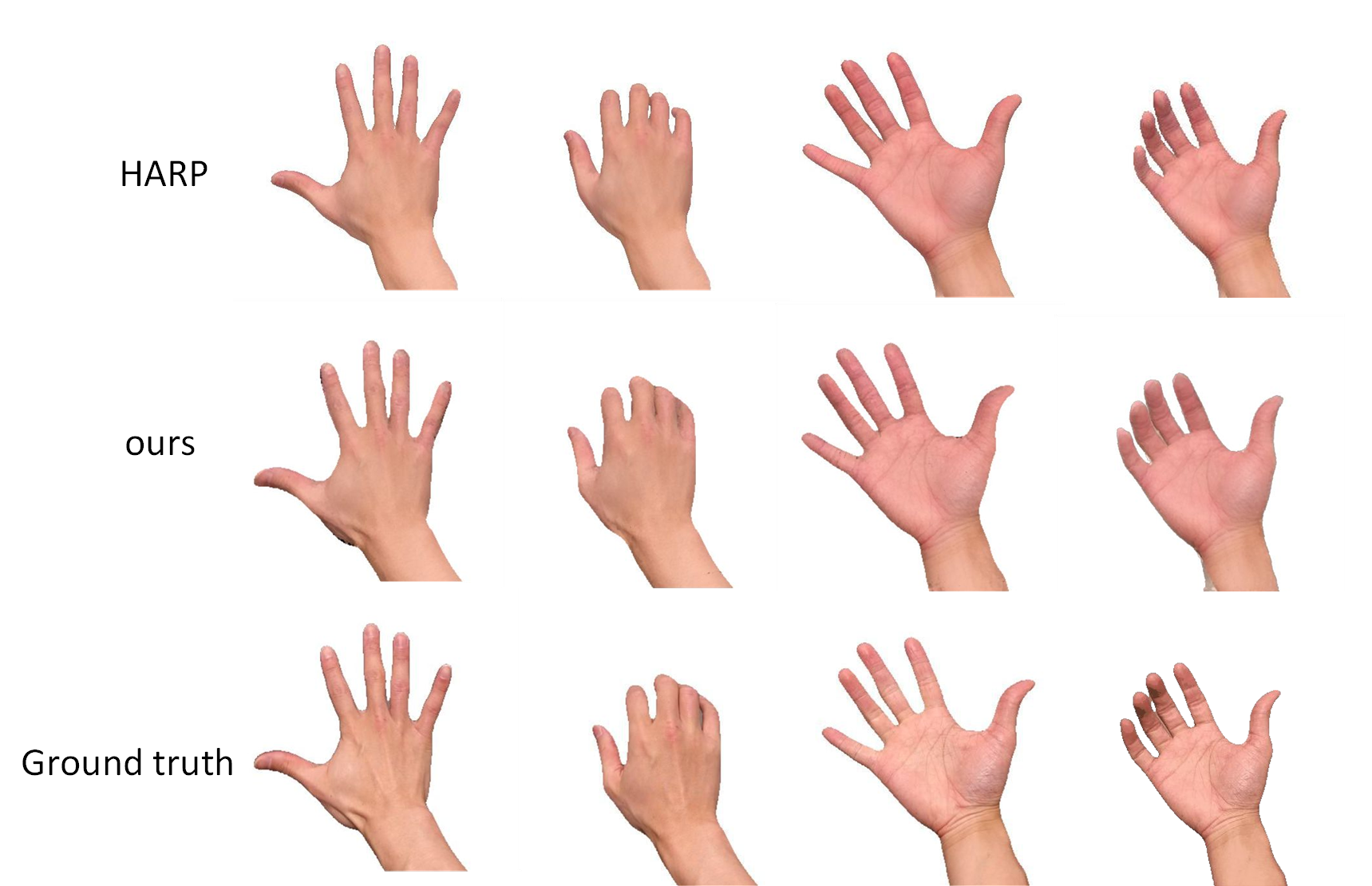}
\end{center}
\vspace{-6mm}
\caption{SOTA Comparison on \textit{Hand Appearance Dataset}}
\label{fig:HARP_comp}
\end{figure}
 \begin{figure}
\begin{center}
\includegraphics[width=7.5cm]{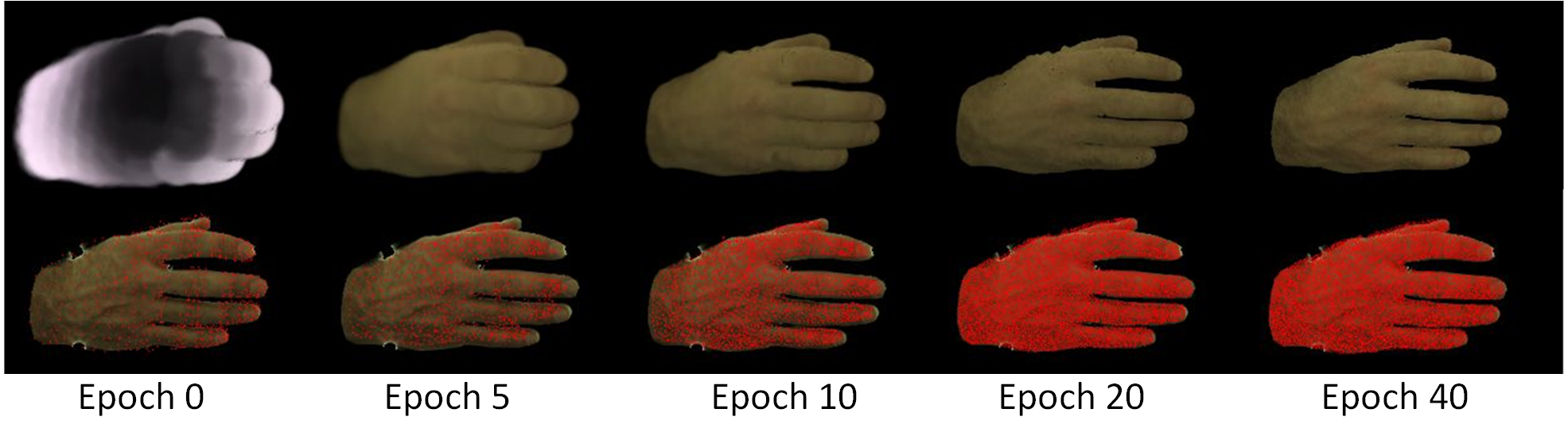}
\end{center}
\vspace{-2mm}
\caption{Optimization of our self-adaptive upsampling strategy. Top: rendered images, bottom: point cloud.}
\vspace{-0.5em}
\label{fig:up_sampling}
\end{figure}

\noindent\textbf{Multi-view}
In this experiment, we train all baseline methods on the multi-view image sequences from the training set of InterHand2.6M\cite{moon2020interhand2}. In $S^2$hand\cite{chen2021model} and AMVUR\cite{jiang2023probabilistic}, input images are required to refine the appearance model of the hand during testing. For fair comparison, we only optimize hand appearance during training. As a result, $S^2$hand\cite{chen2021model} and AMVUR\cite{jiang2023probabilistic} are limited to producing coarse hand texture, as shown in \cref{fig:InterHand_comp}. 
In HTML\cite{qian2020html}, texture parameters of the UV map are optimized using our loss function. While HTML can generate detailed hand texture, it can be inaccurate due to the limitations of its linear appearance model, as shown in \cref{fig:InterHand_comp}. HandAvatar\cite{chen2023hand} employs the volume rendering method to achieve photo-realistic hand reconstruction. However, it requires pre-training the PairOF geometric model. In contrast, our approach involves simultaneous end-to-end training of the geometric and appearance models, resulting in improved geometric fitting and appearance reconstruction, as evidenced by our superior performance, shown in \cref{tab:SOTA_results} and \cref{fig:InterHand_comp}. Further, our method achieves significantly faster training and rendering speeds than NeRF-based approaches, as shown in \cref{section:speed}, and efficiently processes batch images for enhanced training.

\noindent\textbf{Single-view} 
The Hand Appearance dataset \cite{karunratanakul2023harp} is specifically crafted to simulate less constrained capturing conditions, typical for end-users in AR/VR applications. To ensure fair comparison with HARP \cite{karunratanakul2023harp}, we refrained from optimizing any parameters using the testing dataset\tmpdel{, a practice not followed by HARP}. The results presented in \cref{tab:SOTA_results_HARP} and \cref{fig:HARP_comp} highlight our approach's superior performance in both geometric fitting and texture reconstruction for hand-with-arm images.

\begin{table}
\begin{center}
\footnotesize
\caption{Test results on \textit{Hand Appearance Dataset} }
\begin{tabular}{l||p{0.8cm}p{0.8cm}p{0.8cm}p{0.8cm}}
\hline
Method&IOU$\uparrow$&LPIPS$\downarrow$&PSNR$\uparrow$&SSIM$\uparrow$\\
\hline
HARP\cite{karunratanakul2023harp}&0.836&0.128&20.1&0.851\\
ours&0.865&0.117&21.7&0.877\\
\hline
\end{tabular}
\label{tab:SOTA_results_HARP}
\end{center}
\vspace{-1.5em}
\end{table}

\subsection{Ablation Study}
\begin{table}
\footnotesize
\begin{center}
\caption{Impact of each component of our approach\tmpdel{ on rendering quality. All experiments is conducted} on InterHand2.6M \textit{test/Capture0}. Since the second experiment does not account for pose changes, significant improvement is not expected.}
\begin{tabular}{p{0.6cm}p{0.6cm}p{0.6cm}p{0.8cm}|ccc}
\hline
Albedo&Shading&Normal.&Context.&LPIPS$\downarrow$&PSNR$\uparrow$ &$SSIM\uparrow$\\
\hline
 \hfil\checkmark &  &  & &0.106&26.85&0.883\\
\hline
 \hfil\checkmark & \hfil\checkmark &   &&0.105&26.88&0.888\\
\hline
 \hfil\checkmark & \hfil\checkmark & \hfil\checkmark  &  &0.085& 28.65&0.911\\
\hline
 \hfil\checkmark & \hfil\checkmark & \hfil\checkmark  & \hfil\checkmark &0.078 &30.93& 0.934 \\
\hline
\end{tabular}
\label{tab:ablation_rendering}
\end{center}
\vspace{-1.5em}
\end{table}

\begin{figure}
\begin{center}
\includegraphics[width=7.5cm]{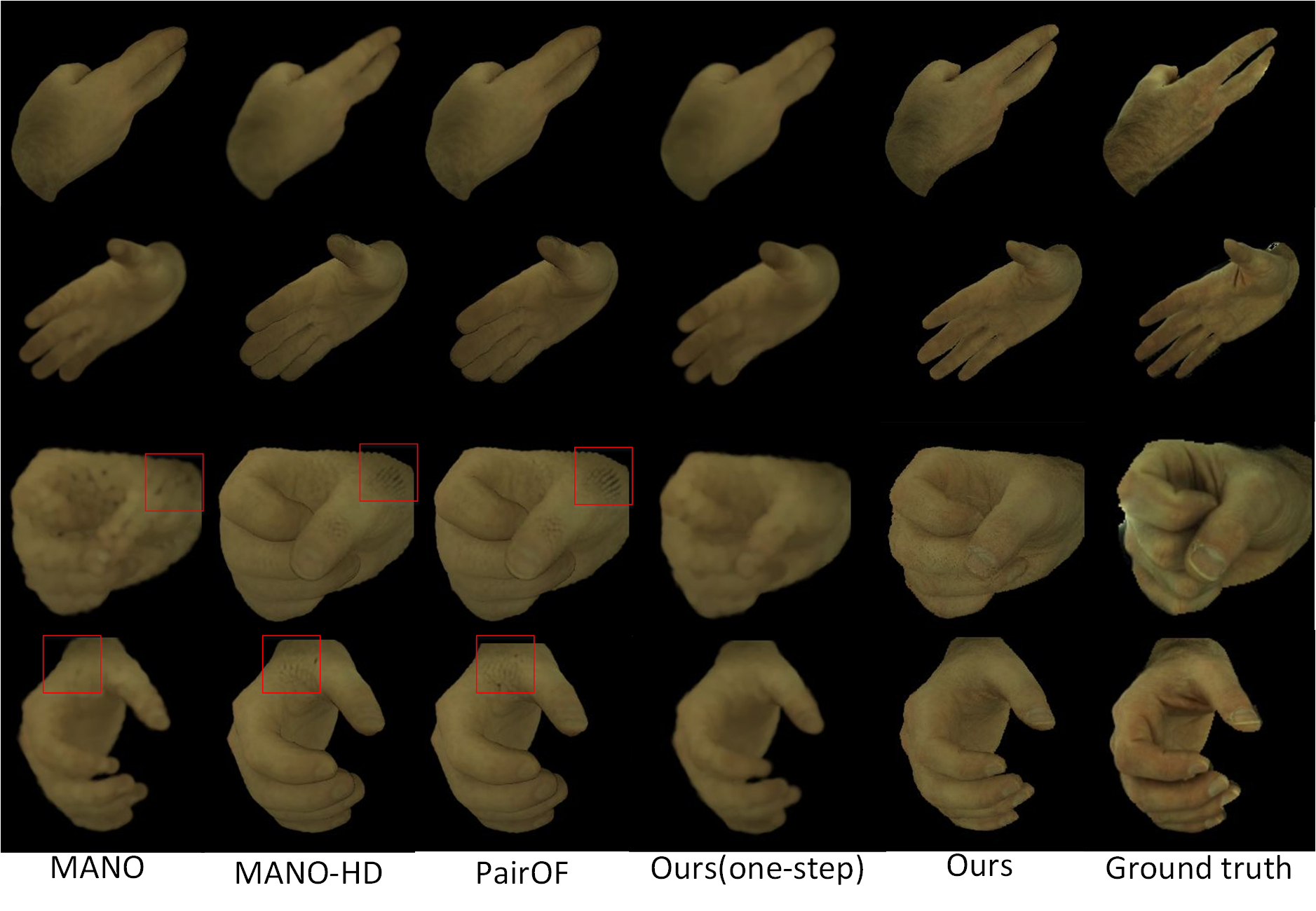}
\end{center}
\vspace{-5mm}
\caption{Geometric representation comparison. MANO, MANO-HD and PairOF contain black holes, highlighted by red rectangle.}
\vspace{-0.5em}
\label{fig:geometry}
\end{figure}

\noindent\textbf{Geometry quality}\label{sec:geometry}
To evaluate the geometry quality of our self-adaptive canonical hand representation, we replace it with other commonly used \tmpdel{hand geometry }representations, including MANO \cite{romero2017embodied}, MANO-HD \cite{chen2023hand} and PairOF \cite{chen2023hand}. MANO-HD is a high-resolution version of MANO achieved by subdividing the template mesh, resulting in 12,337 vertices and 24,608 faces. PairOF is an occupancy field guided by MANO-HD, which requires pre-training\tmpdel{ using all right-hand annotations in InterHand2.6M}. During training on \textit{test/Capture0}, we optimize the coordinates and radius of each baseline method. As shown in \cref{tab:geometry_quality}, our self-adaptive hand geometric model achieves significant improvement, and one-step upsampling significantly impacts performance. As shown in \cref{fig:geometry}, the rendered images of MANO, MANO-HD and PairOF may have black holes due to sparsity or uneven distribution of deformed point clouds, while our self-adaptive hand geometric model shows superior performance, producing smoother and more accurate hand texture. \cref{fig:up_sampling} illustrates the optimization process of our self-adaptive upsampling strategy during training.
\begin{table}
\begin{center}
\footnotesize
\caption{Geometric representation comparison on InterHand2.6M \textit{test/Capture0} for our proposed texture reconstruction model.}
\begin{tabular}{l||p{0.8cm}p{0.8cm}p{0.8cm}p{0.8cm}}
\hline
Geometric representation&IOU$\uparrow$&LPIPS$\downarrow$&PSNR$\uparrow$&SSIM$\uparrow$\\
\hline
MANO\cite{romero2017embodied}&0.887&0.175&25.93&0.851\\
MANO-HD\cite{chen2023hand}&0.863&0.143&25.95&0.866\\
PairOF\cite{chen2023hand}&0.867&0.139&26.10&0.875\\
Ours (one-step upsampling)&0.868&0.172&26.03&0.868\\
Ours&0.946&0.078&30.93&0.934\\
\hline
\end{tabular}
\label{tab:geometry_quality}
\end{center}
\vspace{-1.5em}
\end{table}
 \begin{figure}
\begin{center}
\includegraphics[width=7.5cm]{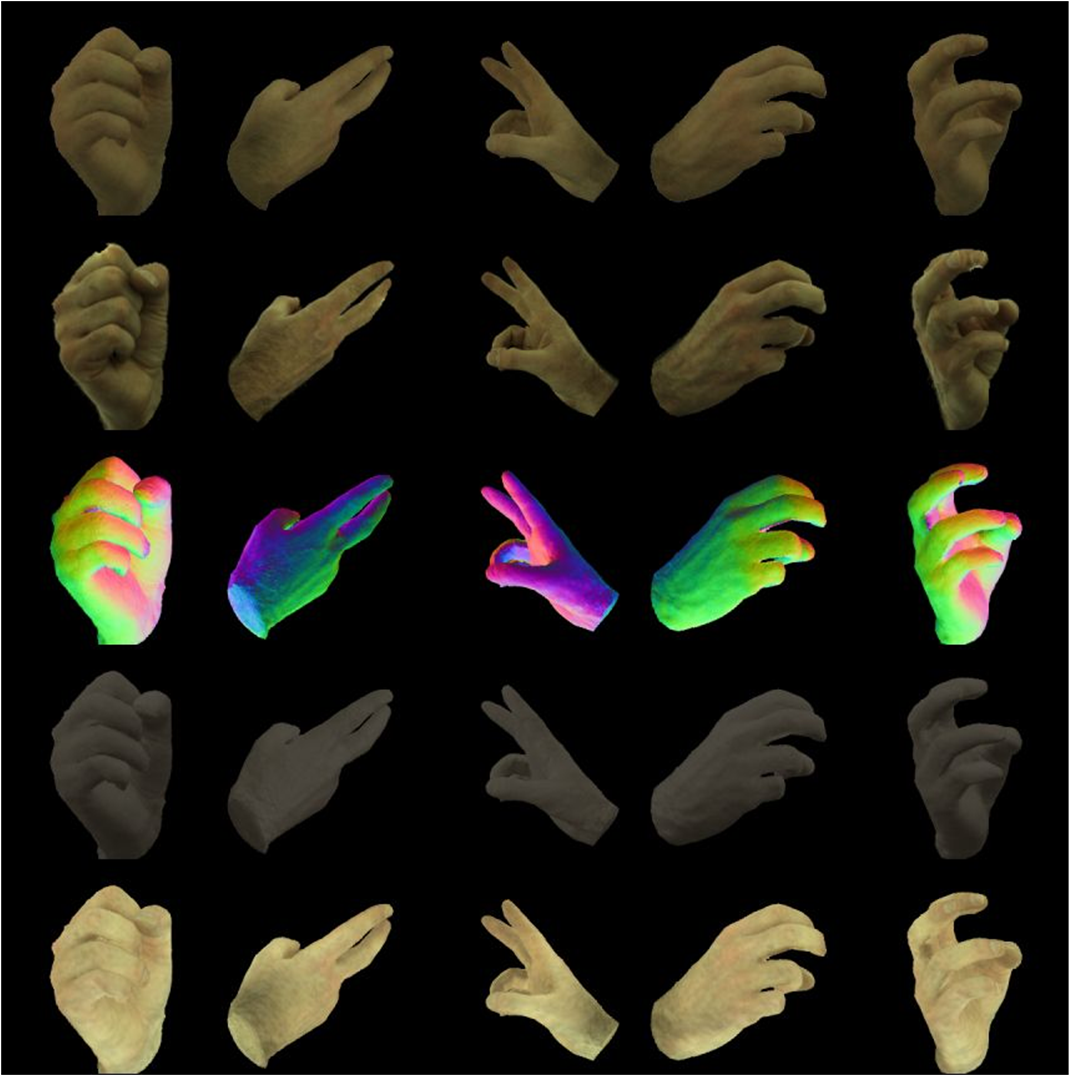}
\end{center}
\vspace{-2mm}
\caption{Output images generated by our approach. Top-bottom: rendered RGB, ground truth, normal, shading and albedo. }
\vspace{-1.5em}
\label{fig:ablation_disentangled}
\end{figure}
\noindent\textbf{Rendering quality:} \tmpdel{In terms of learning albedo and pose-aware shading, a naive approach is to use MLPs. }In \cref{tab:ablation_rendering}, we construct our baseline by only learning albedo with MLPs. In the second experiment, we introduce shading, learned from the canonical normals. Since this does not account for pose changes, significant improvement is not expected. Encoding normal deformation into shading, we observe significant quantitative enhancements\tmpdel{ in rendering metrics, as demonstrated in the third row of \cref{tab:ablation_rendering}. R}, and replacing MLPs with our Context-Attention further improves the \tmpdel{overall} rendering results, highlighting the importance of the proposed Context-Attention module. \cref{fig:ablation_disentangled} shows output images generated by our approach.

 \begin{figure}
\begin{center}
\includegraphics[width=7.5cm]{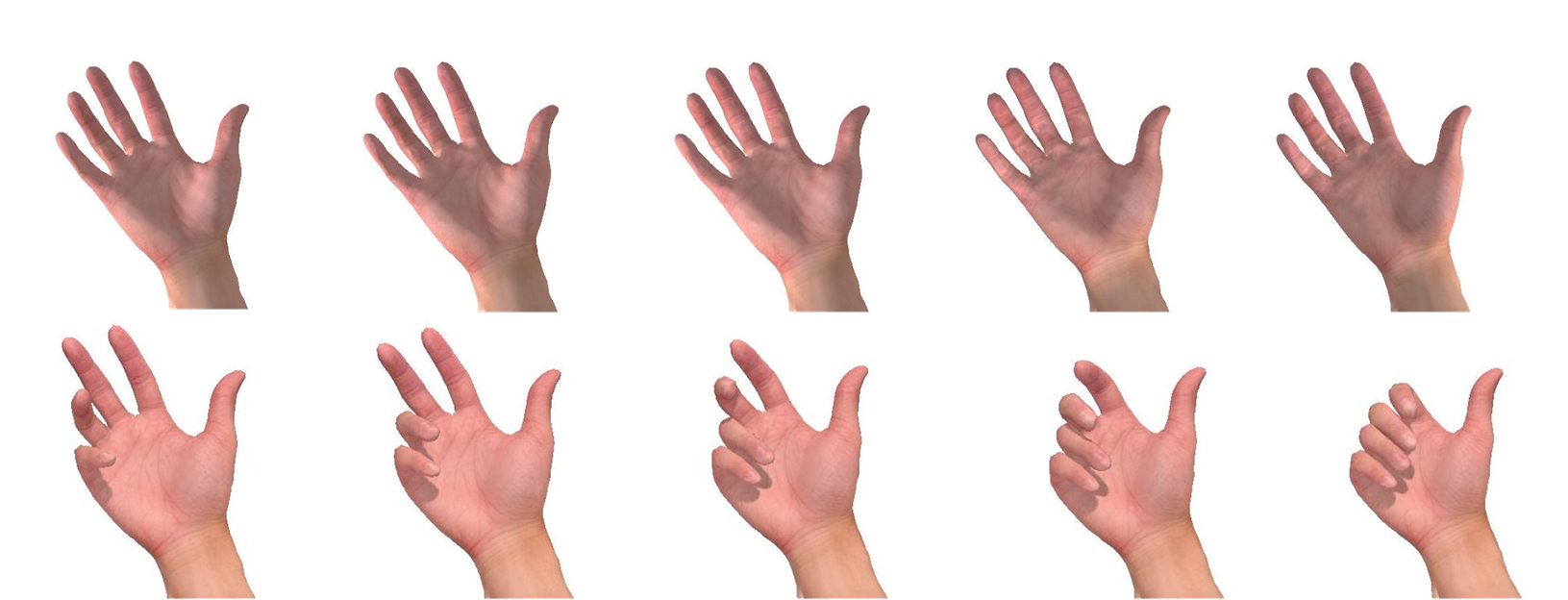}
\end{center}
\vspace{-4mm}
\caption{Generated hand images after relighting. Top: relighted images by adjusting a single lighting source horizontally. Bottom: relighted hand images using self-shadowing.}
\vspace{-0.5em}
\label{fig:lighting}
\end{figure}

\noindent\textbf{Relighting:}\label{sec:relighting} We also demonstrate that our approach can enable relighting by using Phoen model.\tmpdel{ To enable Phong shading using the Phoen model \cite{phong1998illumination}, a hand mesh is required.} In this experiment, we identify the nearest canonical points corresponding to MANO template vertices and utilize these points and the MANO template mesh faces to approximate the hand mesh. We can modify the lighting direction, as shown in \cref{fig:lighting},\tmpdel{. In Figure, we show our approach is also compatible to} and incorporate the self-shadow model \cite{karunratanakul2023harp}\tmpdel{ for estimating self-shadowing between fingers and palm}.

\subsection{Inference speed}\label{section:speed} As shown in Table \ref{tab:speed},  our approach exhibits clear superiority in inference speed compared to other SOTA methods. Although \cite{jiang2023probabilistic} can achieve real-time rendering speed, its reconstruction results are considerably inferior to ours, as shown in \cref{tab:SOTA_results} and \cref{fig:InterHand_comp}. Our method achieves 200 times faster inference speed than the NeRF-based HandAvatar while maintaining superior performance.\tmpdel{ Due to the computation of self-shadowing, the relighting version of our approach is slightly slower than our original implementation, but it still achieves real-time inference speed.}

\begin{table}
\small
\begin{center}
\caption{Inference time (seconds/frame) on Inter-Hand2.6M test/Capture0 testing.}
\begin{tabular}{p{0.6cm}|p{1.4cm}|p{1.1cm}|p{0.8cm}|p{0.6cm}|p{1.2cm}}
\hline
 &HandAvatar\cite{chen2023hand} &AMVUR\cite{jiang2023probabilistic} &HARP\cite{karunratanakul2023harp}&ours&ours (relighting)\\
\hline
Time &3.77 &0.021&0.057&0.018&0.023\\
\hline
\end{tabular}
\label{tab:speed}
\end{center}
\vspace{-1.5em}
\end{table}

\section{Conclusion}
We propose 3D-PSHR, a novel real-time and photo-realistic hand reconstruction approach that represents the hand using deformable point clouds and achieves rendering through 3D Point Splatting. Our approach incorporates a self-adaptive canonical points upsampling strategy to dynamically enhance the resolution of the hand's topology structure, capturing finer texture details. Additionally, we propose a self-adaptive deformation approach to deform canonical points to the target pose. To model hand texture, we leverage intrinsic albedo and pose-aware shading learned through Context-Attention. Notably, our method is trainable on pose-free and view-free hand datasets. It demonstrates robust performance when trained with monocular or multi-view videos, achieving real-time rendering speed with superior results. Moreover, our approach supports relighting with the Phoen or self-shadow models.
\tmpdel{ In the future, we will explore optimizing our self-adaptive canonical points upsampling strategy to achieve a satisfactory hand geometry representation with a minimal number of points, thus improving training and inference speed.}
{\small
\bibliographystyle{ieeetr}
\bibliography{egbib}

\begin{thebibliography}{10}

\bibitem{cao2022authentic}
C.~Cao, T.~Simon, J.~K. Kim, G.~Schwartz, M.~Zollhoefer, S.-S. Saito, S.~Lombardi, S.-E. Wei, D.~Belko, S.-I. Yu, {\em et~al.}, ``Authentic volumetric avatars from a phone scan,'' {\em ACM Transactions on Graphics (TOG)}, vol.~41, no.~4, pp.~1--19, 2022.

\bibitem{grassal2022neural}
P.-W. Grassal, M.~Prinzler, T.~Leistner, C.~Rother, M.~Nie{\ss}ner, and J.~Thies, ``Neural head avatars from monocular rgb videos,'' in {\em Proceedings of the IEEE/CVF Conference on Computer Vision and Pattern Recognition}, pp.~18653--18664, 2022.

\bibitem{jiang2022selfrecon}
B.~Jiang, Y.~Hong, H.~Bao, and J.~Zhang, ``Selfrecon: Self reconstruction your digital avatar from monocular video,'' in {\em Proceedings of the IEEE/CVF Conference on Computer Vision and Pattern Recognition}, pp.~5605--5615, 2022.

\bibitem{weng2022humannerf}
C.-Y. Weng, B.~Curless, P.~P. Srinivasan, J.~T. Barron, and I.~Kemelmacher-Shlizerman, ``Humannerf: Free-viewpoint rendering of moving people from monocular video,'' in {\em Proceedings of the IEEE/CVF conference on computer vision and pattern Recognition}, pp.~16210--16220, 2022.

\bibitem{zheng2023pointavatar}
Y.~Zheng, W.~Yifan, G.~Wetzstein, M.~J. Black, and O.~Hilliges, ``Pointavatar: Deformable point-based head avatars from videos,'' in {\em Proceedings of the IEEE/CVF Conference on Computer Vision and Pattern Recognition}, pp.~21057--21067, 2023.

\bibitem{qian2020html}
N.~Qian, J.~Wang, F.~Mueller, F.~Bernard, V.~Golyanik, and C.~Theobalt, ``Html: A parametric hand texture model for 3d hand reconstruction and personalization,'' in {\em Computer Vision--ECCV 2020: 16th European Conference, Glasgow, UK, August 23--28, 2020, Proceedings, Part XI 16}, pp.~54--71, Springer, 2020.

\bibitem{kopanas2022neural}
G.~Kopanas, T.~Leimk{\"u}hler, G.~Rainer, C.~Jambon, and G.~Drettakis, ``Neural point catacaustics for novel-view synthesis of reflections,'' {\em ACM Transactions on Graphics (TOG)}, vol.~41, no.~6, pp.~1--15, 2022.

\bibitem{karunratanakul2023harp}
K.~Karunratanakul, S.~Prokudin, O.~Hilliges, and S.~Tang, ``Harp: Personalized hand reconstruction from a monocular rgb video,'' in {\em Proceedings of the IEEE/CVF Conference on Computer Vision and Pattern Recognition}, pp.~12802--12813, 2023.

\bibitem{chen2021model}
Y.~Chen, Z.~Tu, D.~Kang, L.~Bao, Y.~Zhang, X.~Zhe, R.~Chen, and J.~Yuan, ``Model-based 3d hand reconstruction via self-supervised learning,'' in {\em Proceedings of the IEEE/CVF Conference on Computer Vision and Pattern Recognition}, pp.~10451--10460, 2021.

\bibitem{jiang2023probabilistic}
Z.~Jiang, H.~Rahmani, S.~Black, and B.~M. Williams, ``A probabilistic attention model with occlusion-aware texture regression for 3d hand reconstruction from a single rgb image,'' in {\em Proceedings of the IEEE/CVF Conference on Computer Vision and Pattern Recognition}, pp.~758--767, 2023.

\bibitem{chen2023hand}
X.~Chen, B.~Wang, and H.-Y. Shum, ``Hand avatar: Free-pose hand animation and rendering from monocular video,'' in {\em Proceedings of the IEEE/CVF Conference on Computer Vision and Pattern Recognition}, pp.~8683--8693, 2023.

\bibitem{guo2023handnerf}
Z.~Guo, W.~Zhou, M.~Wang, L.~Li, and H.~Li, ``Handnerf: Neural radiance fields for animatable interacting hands,'' in {\em Proceedings of the IEEE/CVF Conference on Computer Vision and Pattern Recognition}, pp.~21078--21087, 2023.

\bibitem{corona2022lisa}
E.~Corona, T.~Hodan, M.~Vo, F.~Moreno-Noguer, C.~Sweeney, R.~Newcombe, and L.~Ma, ``Lisa: Learning implicit shape and appearance of hands,'' in {\em Proceedings of the IEEE/CVF Conference on Computer Vision and Pattern Recognition}, pp.~20533--20543, 2022.

\bibitem{mildenhall2021nerf}
B.~Mildenhall, P.~P. Srinivasan, M.~Tancik, J.~T. Barron, R.~Ramamoorthi, and R.~Ng, ``Nerf: Representing scenes as neural radiance fields for view synthesis,'' {\em Communications of the ACM}, vol.~65, no.~1, pp.~99--106, 2021.

\bibitem{li2022nimble}
Y.~Li, L.~Zhang, Z.~Qiu, Y.~Jiang, N.~Li, Y.~Ma, Y.~Zhang, L.~Xu, and J.~Yu, ``Nimble: a non-rigid hand model with bones and muscles,'' {\em ACM Transactions on Graphics (TOG)}, vol.~41, no.~4, pp.~1--16, 2022.

\bibitem{romero2017embodied}
J.~ROMERO, D.~TZIONAS, and M.~J. BLACK, ``Embodied hands: Modeling and capturing hands and bodies together** supplementary material,'' 2017.

\bibitem{alldieck2021imghum}
T.~Alldieck, H.~Xu, and C.~Sminchisescu, ``imghum: Implicit generative models of 3d human shape and articulated pose,'' in {\em Proceedings of the IEEE/CVF International Conference on Computer Vision}, pp.~5461--5470, 2021.

\bibitem{mihajlovic2021leap}
M.~Mihajlovic, Y.~Zhang, M.~J. Black, and S.~Tang, ``Leap: Learning articulated occupancy of people,'' in {\em Proceedings of the IEEE/CVF Conference on Computer Vision and Pattern Recognition}, pp.~10461--10471, 2021.

\bibitem{karunratanakul2020grasping}
K.~Karunratanakul, J.~Yang, Y.~Zhang, M.~J. Black, K.~Muandet, and S.~Tang, ``Grasping field: Learning implicit representations for human grasps,'' in {\em 2020 International Conference on 3D Vision (3DV)}, pp.~333--344, IEEE, 2020.

\bibitem{zheng2022imface}
M.~Zheng, H.~Yang, D.~Huang, and L.~Chen, ``Imface: A nonlinear 3d morphable face model with implicit neural representations,'' in {\em Proceedings of the IEEE/CVF Conference on Computer Vision and Pattern Recognition}, pp.~20343--20352, 2022.

\bibitem{deng2022gram}
Y.~Deng, J.~Yang, J.~Xiang, and X.~Tong, ``Gram: Generative radiance manifolds for 3d-aware image generation,'' in {\em Proceedings of the IEEE/CVF Conference on Computer Vision and Pattern Recognition}, pp.~10673--10683, 2022.

\bibitem{jiang2022neuman}
W.~Jiang, K.~M. Yi, G.~Samei, O.~Tuzel, and A.~Ranjan, ``Neuman: Neural human radiance field from a single video,'' in {\em European Conference on Computer Vision}, pp.~402--418, Springer, 2022.

\bibitem{liu2021neural}
L.~Liu, M.~Habermann, V.~Rudnev, K.~Sarkar, J.~Gu, and C.~Theobalt, ``Neural actor: Neural free-view synthesis of human actors with pose control,'' {\em ACM transactions on graphics (TOG)}, vol.~40, no.~6, pp.~1--16, 2021.

\bibitem{qiao2023dynamic}
Y.-L. Qiao, A.~Gao, Y.~Xu, Y.~Feng, J.-B. Huang, and M.~C. Lin, ``Dynamic mesh-aware radiance fields,'' in {\em Proceedings of the IEEE/CVF International Conference on Computer Vision}, pp.~385--396, 2023.

\bibitem{xu2022point}
Q.~Xu, Z.~Xu, J.~Philip, S.~Bi, Z.~Shu, K.~Sunkavalli, and U.~Neumann, ``Point-nerf: Point-based neural radiance fields,'' in {\em Proceedings of the IEEE/CVF Conference on Computer Vision and Pattern Recognition}, pp.~5438--5448, 2022.

\bibitem{grossman1998point}
J.~P. Grossman and W.~J. Dally, ``Point sample rendering,'' in {\em Rendering Techniques’ 98: Proceedings of the Eurographics Workshop in Vienna, Austria, June 29—July 1, 1998 9}, pp.~181--192, Springer, 1998.

\bibitem{botsch2005high}
M.~Botsch, A.~Hornung, M.~Zwicker, and L.~Kobbelt, ``High-quality surface splatting on today's gpus,'' in {\em Proceedings Eurographics/IEEE VGTC Symposium Point-Based Graphics, 2005.}, pp.~17--141, IEEE, 2005.

\bibitem{pfister2000surfels}
H.~Pfister, M.~Zwicker, J.~Van~Baar, and M.~Gross, ``Surfels: Surface elements as rendering primitives,'' in {\em Proceedings of the 27th annual conference on Computer graphics and interactive techniques}, pp.~335--342, 2000.

\bibitem{ren2002object}
L.~Ren, H.~Pfister, and M.~Zwicker, ``Object space ewa surface splatting: A hardware accelerated approach to high quality point rendering,'' in {\em Computer Graphics Forum}, vol.~21, pp.~461--470, Wiley Online Library, 2002.

\bibitem{zwicker2001surface}
M.~Zwicker, H.~Pfister, J.~Van~Baar, and M.~Gross, ``Surface splatting,'' in {\em Proceedings of the 28th annual conference on Computer graphics and interactive techniques}, pp.~371--378, 2001.

\bibitem{wiles2020synsin}
O.~Wiles, G.~Gkioxari, R.~Szeliski, and J.~Johnson, ``Synsin: End-to-end view synthesis from a single image,'' in {\em Proceedings of the IEEE/CVF Conference on Computer Vision and Pattern Recognition}, pp.~7467--7477, 2020.

\bibitem{yifan2019differentiable}
W.~Yifan, F.~Serena, S.~Wu, C.~{\"O}ztireli, and O.~Sorkine-Hornung, ``Differentiable surface splatting for point-based geometry processing,'' {\em ACM Transactions on Graphics (TOG)}, vol.~38, no.~6, pp.~1--14, 2019.

\bibitem{kerbl20233d}
B.~Kerbl, G.~Kopanas, T.~Leimk{\"u}hler, and G.~Drettakis, ``3d gaussian splatting for real-time radiance field rendering,'' {\em ACM Transactions on Graphics (ToG)}, vol.~42, no.~4, pp.~1--14, 2023.

\bibitem{gropp2020implicit}
A.~Gropp, L.~Yariv, N.~Haim, M.~Atzmon, and Y.~Lipman, ``Implicit geometric regularization for learning shapes,'' {\em arXiv preprint arXiv:2002.10099}, 2020.

\bibitem{romero2022embodied}
J.~Romero, D.~Tzionas, and M.~J. Black, ``Embodied hands: Modeling and capturing hands and bodies together,'' {\em arXiv preprint arXiv:2201.02610}, 2022.

\bibitem{phong1998illumination}
B.~T. Phong, ``Illumination for computer generated pictures,'' in {\em Seminal graphics: pioneering efforts that shaped the field}, pp.~95--101, 1998.

\bibitem{ledig2017photo}
C.~Ledig, L.~Theis, F.~Husz{\'a}r, J.~Caballero, A.~Cunningham, A.~Acosta, A.~Aitken, A.~Tejani, J.~Totz, Z.~Wang, {\em et~al.}, ``Photo-realistic single image super-resolution using a generative adversarial network,'' in {\em Proceedings of the IEEE conference on computer vision and pattern recognition}, pp.~4681--4690, 2017.

\bibitem{moon2020interhand2}
G.~Moon, S.-I. Yu, H.~Wen, T.~Shiratori, and K.~M. Lee, ``Interhand2. 6m: A dataset and baseline for 3d interacting hand pose estimation from a single rgb image,'' in {\em Computer Vision--ECCV 2020: 16th European Conference, Glasgow, UK, August 23--28, 2020, Proceedings, Part XX 16}, pp.~548--564, Springer, 2020.

\bibitem{paszke2019pytorch}
A.~Paszke, S.~Gross, F.~Massa, A.~Lerer, J.~Bradbury, G.~Chanan, T.~Killeen, Z.~Lin, N.~Gimelshein, L.~Antiga, {\em et~al.}, ``Pytorch: An imperative style, high-performance deep learning library,'' {\em Advances in neural information processing systems}, vol.~32, 2019.

\bibitem{kingma2014adam}
D.~P. Kingma and J.~Ba, ``Adam: A method for stochastic optimization,'' {\em arXiv preprint arXiv:1412.6980}, 2014.

\end{thebibliography}
}
\tmpdel{
\supp{\newpage
\section{Supplementary}
For optimizing
the our model, we use $\lambda_{rgb} = 1$, $\lambda_{mask} = 1$, $\lambda_{vgg} = 0.1$, $\lambda_{reg} = 1$, $\lambda_{SDF} = 1$ and $\lambda_{eik} = 0.1$. 
\begin{figure}
\begin{center}
\includegraphics[width=8cm]{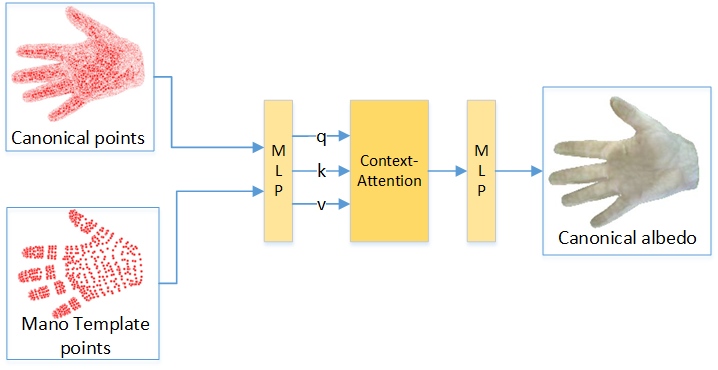}
\end{center}
\vspace{-5mm}
\caption{Context-attention module for learning canonical albedo}
\label{fig:context_attention}
\end{figure}}

}
\end{document}


\title{3D Points Splatting for Real-Time Dynamic Hand Reconstruction (Supplementary Material)}

\author{First Author\\
Institution1\\
Institution1 address\\
{\tt\small firstauthor@i1.org}
\and
Second Author\\
Institution2\\
First line of institution2 address\\
{\tt\small secondauthor@i2.org}
}

\maketitle

\section{Context-Attention Module }
\begin{figure}
\begin{center}
\includegraphics[width=7.5cm]{images/cross_attention.png}\\ \includegraphics[width=7.5cm]{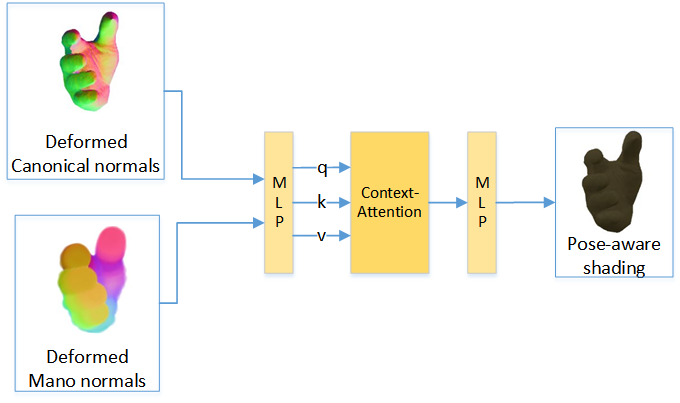}
\end{center}
\vspace{-5mm}
\caption{Context-attention modules for learning canonical albedo (Top) and shading (Bottom)}
\label{fig:context_attention}
\end{figure}

The Context-attention modules for learning canonical albedo and shading, introduced in Section 3.3 of the main paper, are shown in \cref{fig:context_attention}. The canonical albedo module takes the 3D co-ordinates of the canonical and MANO template points as input to produce the canonical albedo. The shading module takes the deformed canonical normals and deformed MANO normals as input to produce the pose-aware shading result.

\section{Ablation }
\subsection{Pose parameter and appearance optimization}
\begin{figure}
\begin{center}
\includegraphics[width=8cm]{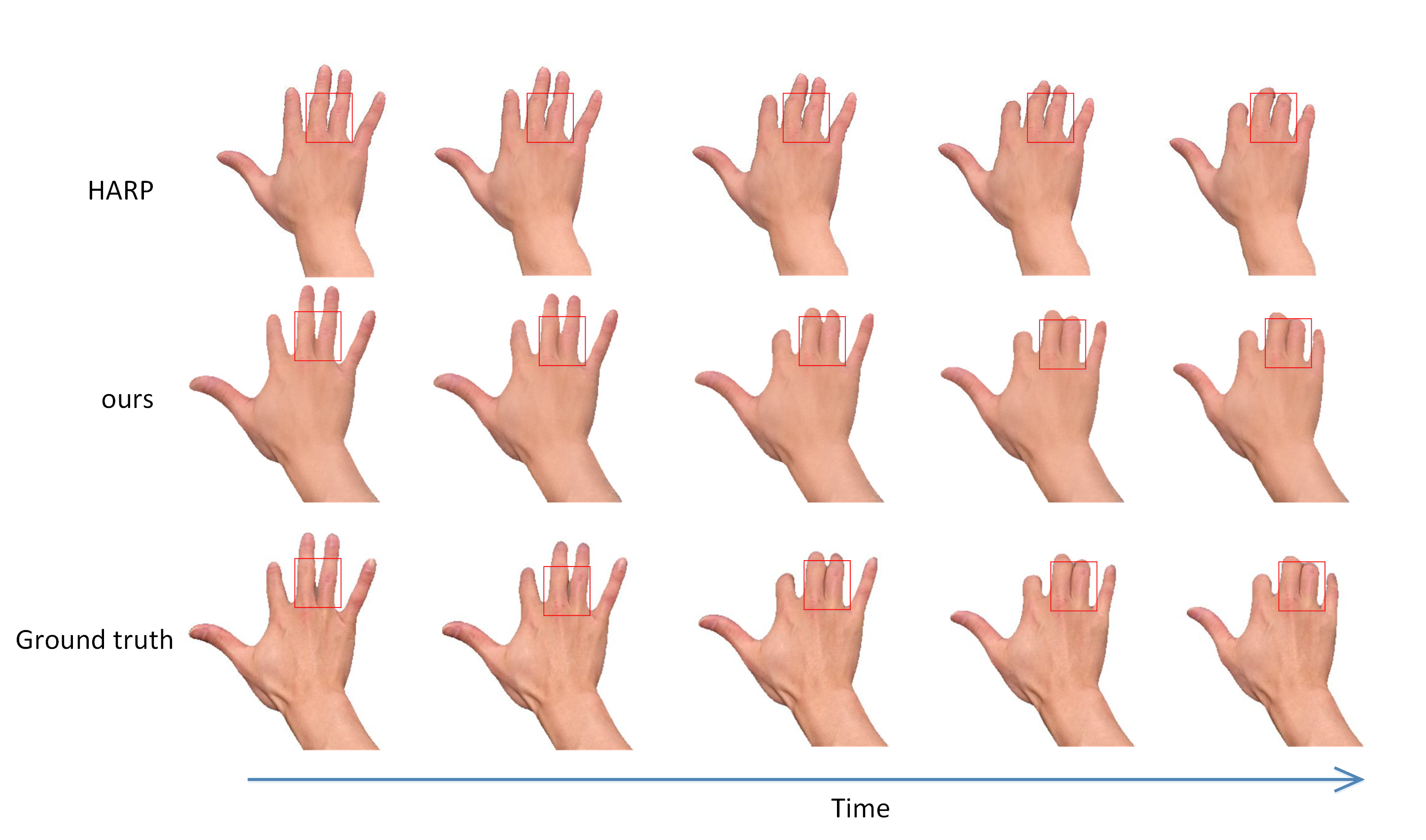}
\end{center}
\vspace{-5mm}
\caption{SOTA Comparison on Hand Appearance Dataset}
\vspace{-1.5em}
\label{fig:supp_harp}
\end{figure}

In the work presented by \cite{karunratanakul2023harp}, the authors optimize the pose parameters of the MANO model and further refine the appearance based on testing images. We follow their experimental setting on the \textit{Hand Appearance Dataset} and report the results in \cref{tab:HARP_test}. Several examples are shown in \cref{fig:supp_harp}, demonstrating superior geometry fitting and appearance reconstruction of our approach.
\begin{table}
\begin{center}
\footnotesize
\caption{Test results on \textit{Hand Appearance Dataset} }
\begin{tabular}{l||p{0.8cm}p{0.8cm}p{0.8cm}p{0.8cm}}
\hline
Method&IOU$\uparrow$&LPIPS$\downarrow$&PSNR$\uparrow$&SSIM$\uparrow$\\
\hline
HARP\cite{karunratanakul2023harp}&0.861&0.107&21.50&0.878\\
ours&0.901&0.095&23.55&0.893\\
\hline
\end{tabular}
\label{tab:HARP_test}
\end{center}
\vspace{-1.5em}
\end{table}

\subsection{Effect of loss weight}
We group our loss weights into appearance ($\lambda_{rgb}$,$\lambda_{vgg}$), geometry ($\lambda_{reg}$,$\lambda_{SDF}$,$\lambda{eik}$) and silhouette loss weights ($\lambda_{mask}$) and carry out ablation studies on each group while keeping the parameters of the remaining groups fixed at their determined optimal values.

\paragraph{Appearance loss weights:} Quantitative comparison of appearance loss weights is shown in \cref{tab:appearance_loss_weights}. Since we have two terms in this loss, both with coefficients, we fix $\lambda_{rgb}$ to 1 for most experiments. Geometry and Silhouette loss are kept with the optimal parameters for all experiments. We observe clear degradation in the performance of our approach when either the RGB or VGG loss are not used, which is significant when $\lambda_{rgb}=0$ or $\lambda_{vgg}$ is very high. Our approach achieves the best performance when configured with $\lambda_{rgb}=1$ and $\lambda_{vgg}=0.1$.

\begin{figure}
\begin{center}
\includegraphics[width=8cm]{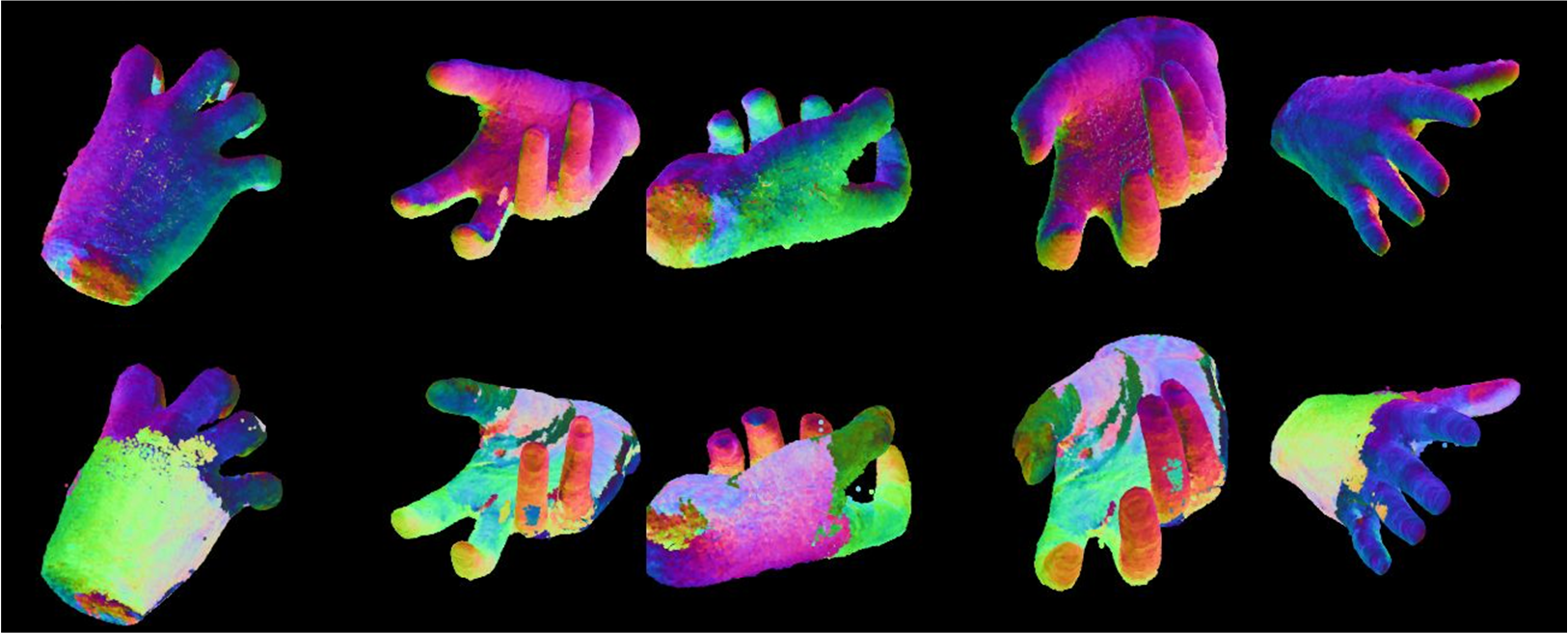}
\end{center}
\caption{The smooth signed distance function improve the surface smoothness. The top row displays hand normal images with the experimental settings $\lambda_{reg}=1,\lambda_{SDF}=1,\lambda_{eik}=0.1$. The bottom row shows hand normal images with the experimental setting $\lambda_{reg}=0$}
\label{fig:smooth}
\end{figure}

\paragraph{Geometry loss weights:}
Quantitative comparison of geometry loss weights is shown in \cref{tab:geometry_loss_weights}. Similiar to the above, we set $\lambda_{SDF}=1$ for all experiments and search for the optimal $\lambda_{reg}$ and $\lambda_{eik}$.
Setting $\lambda_{reg}=0$ is equivalent to setting $\lambda_{SDF}$ and $\lambda_{eik}$ to zero. The optimal parameters are determined to be $(\lambda_{reg},\lambda_{SDF},\lambda_{eik}=(1,1,0.1)$. Qualitiative results are shown in \cref{fig:smooth}, demonstrating the impact of the geometry loss function. It can be clearly seen that excluding this loss results in considerable noise in the normal images, while using our optimal parameters produces smooth realistic normals.

\paragraph{Silhouette loss weight:}
Quantitative comparison of silhouette loss weight is shown in \cref{tab:silhouette_loss_weights}. Our approach achieves the best performance when configured with $\lambda_{mask}=1$.

\begin{table}
\begin{center}
\footnotesize
\caption{Ablation study on the appearance loss weights}
\begin{tabular}{p{0.8cm}p{0.8cm}||p{0.8cm}p{0.8cm}p{0.8cm}p{0.8cm}}
\hline
$\lambda_{rgb}$&$\lambda_{vgg}$&IOU$\uparrow$&LPIPS$\downarrow$&PSNR$\uparrow$&SSIM$\uparrow$\\
\hline
0&1&0.944&0.133&21.97&0.880\\
1&0&0.943&0.088&29.87&0.928\\
1&10&0.943&0.118&22.81&0.893\\
1&0.2&0.945&0.084&30.34&0.932\\
1&0.1&\textbf{0.946} &\textbf{0.078} &\textbf{30.93} &\textbf{0.934}\\
1&0.05&0.945&0.080 &30.54 &0.934 \\
\end{tabular}
\label{tab:appearance_loss_weights}
\end{center}
\vspace{-1.5em}
\end{table}

\begin{table}
\begin{center}
\footnotesize
\caption{Ablation study on the geometry loss weights}
\begin{tabular}{p{0.6cm}p{0.6cm}p{0.6cm}||p{0.8cm}p{0.8cm}p{0.8cm}p{0.8cm}}
\hline
$\lambda_{reg}$&$\lambda_{SDF}$&$\lambda_{eik}$&IOU$\uparrow$&LPIPS$\downarrow$&PSNR$\uparrow$&SSIM$\uparrow$\\
\hline
1&1&10&0.934&0.088&29.98&0.928\\
1&1&1&0.941&0.081&30.72&0.932\\
1&1&0.2&0.942&0.081&30.77&0.932\\
1&1&0.1&\textbf{0.946} &\textbf{0.078} &\textbf{30.93} &\textbf{0.934}\\
1&1&0.05& 0.943&0.080 &30.82 &0.932 \\
1&1&0.01&0.939&0.081&30.63&0.930\\
\hline
0.5&1&0.1&0.942 &0.080&30.74 & 0.931\\
\hline
2&1&0.1&0.938&0.082&30.60&0.931\\
\hline
0&1&0.1&0.934&0.084&30.15&0.928\\
\end{tabular}
\label{tab:geometry_loss_weights}
\end{center}
\vspace{-1.5em}
\end{table}

\begin{table}
\begin{center}
\footnotesize
\caption{Ablation study on the silhouette loss weight}
\begin{tabular}{l||p{0.8cm}p{0.8cm}p{0.8cm}p{0.8cm}}
\hline
&IOU$\uparrow$&LPIPS$\downarrow$&PSNR$\uparrow$&SSIM$\uparrow$\\
\hline
$\lambda_{mask}=0$&0.938&0.083&30.19&0.930\\
$\lambda_{mask}=0.1$&0.941&0.080&30.78&0.931\\
$\lambda_{mask}=0.5$&0.943&0.080&30.90&0.933\\
$\lambda_{mask}=1$&\textbf{0.946}& \textbf{0.078}& \textbf{30.93}& \textbf{0.934}\\
$\lambda_{mask}=2$&0.945&0.081&30.83&0.930\\
$\lambda_{mask}=10$&0.946&0.083&30.75&0.927\\
\hline
\end{tabular}
\label{tab:silhouette_loss_weights}
\end{center}
\vspace{-1.5em}
\end{table}

\section{Training algorithm}
\begin{algorithm}[H]
\small
\caption{Training algorithm}
\label{algorithm}
\begin{algorithmic}[1]
\renewcommand{\algorithmicrequire}{\textbf{Input:} }
\REQUIRE MANO pose $\theta$ and shape $\phi$ parameters; Camera parameters;
\STATE Initialize 3D coordinates canonical points ($P_{C}$) with the MANO template. 
\FOR {epoch $e\leq E$ }
    \IF {$e\%5==0$ and $e<=35$}
        \STATE Upsample canonical points and reduce radius by a factor of $1/\sqrt{2}$
    \ENDIF
    
    \STATE Set the visibility of all canonical points to False
     \FOR {each image batch }
        \STATE Compute $\mathcal{F}_{SDF}(p_{C}^{i})$ and $\bigtriangledown_{p}\mathcal{F}_{SDF}(p)$, where $p_{C}^{i} \in \{P_{C}$ and $p\in \left\{P_{C}, \Omega\right\}$ for smoothing hand surface constructed from our canonical points
        \STATE Compute Eq.(2) to obtain deformed canonical points.
        \STATE Compute Eq.(3) to obtain albedo colors of canonical points
        \STATE Compute Eq.(4-6) to obtain shading values of deformed canonical points
        \STATE Compute the color of each deformed canonical points by the element-wise product between the albedo colors and shading values
        \STATE Compute Eq.(7) to obtain the rendered hand image.
        \STATE Calculate loss by using Eq.(8) and optimize our model
        \STATE Project MANO mesh to obtain the foreground boundary of the hand.
        \STATE Set all points that fall inside the foreground boundary of the hand to True
    \ENDFOR
    \IF {$e<=35$}
        \STATE Prune invisible canonical points
    \ELSE
        \STATE Fix 3D coordinates and the number of canonical points
    \ENDIF
        
\ENDFOR
\end{algorithmic}
\end{algorithm}

\section{Comparision with SOTA}
\begin{figure*}
\begin{center}
\includegraphics[width=15cm]{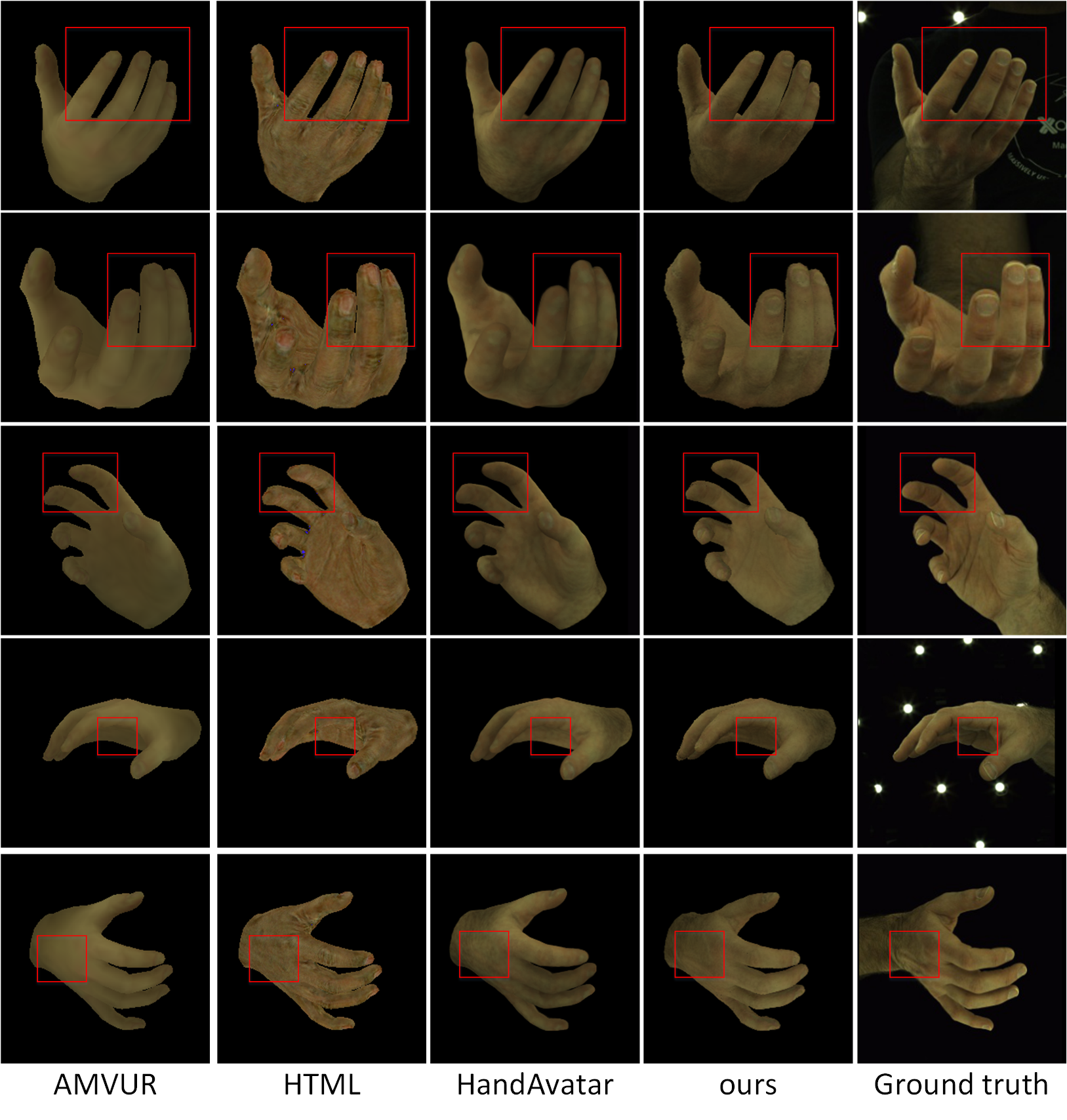}
\end{center}
\vspace{-6mm}
\caption{Comparison with SOTA Methods on InterHand2.6M.  }
\vspace{-1.5em}
\label{fig:examples_supp}
\end{figure*}

Additional examples for comparison with SOTA methods are shown in \cref{fig:examples_supp}

{\small
\bibliographystyle{ieeetr}
\bibliography{egbib}
}